\pdfoutput=1
\documentclass[letterpaper, 10 pt, conference]{ieeeconf}  % Comment this line out if you need a4paper

\IEEEoverridecommandlockouts                              % This command is only needed if 
\usepackage{amsmath}
{}
{}
{}
\usepackage{caption}
\usepackage{subcaption}
\usepackage[acronym]{glossaries}
\glsdisablehyper

\usepackage{hyperref}
\usepackage{booktabs}

\usepackage{pgfplots}
\usepackage{tikz}
\usetikzlibrary{positioning}

\usepackage{color, colortbl}
\definecolor{Gray}{gray}{0.9}

\usepackage{longtable}

\usepackage{array}

\let\labelindent\relax
\usepackage{enumitem}

\usepackage[tikz]{ocgx2}
\usepackage{pgfplotstable}
\pgfplotsset{width=12.2cm, height=7cm, compat=1.14}
\usepackage{svg}

\usepackage[font={footnotesize}]{caption}

\usepackage{amssymb}
\usepackage{booktabs,xltabular}

\usepackage{ amssymb }
\usepackage{siunitx}
\usepackage{ulem}

\newacronym{adas}{ADAS}{advanced driver-assistance systems}
\newacronym{hsv}{HSV}{Hue, Saturation, Value}
\newacronym{iou}{IoU}{Intersection Over Union}
\newacronym{giou}{GIoU}{Generalized Intersection over Union}
\newacronym{diou}{DIoU}{Distance-IoU}
\newacronym{ciou}{CIoU}{Complete IoU}
\newacronym{bcewll}{BCEWithLogitsLoss}{binary cross-entropy with logits loss}
\newacronym{ipm}{IPM}{Inverse Perspective Mapping}
\newacronym{yolo}{YOLO}{You Only Look Once}
\newacronym{bev}{BEV}{Bird's Eye View}
\newacronym{dnn}{DNN}{Deep Neural Network}
\newacronym{cnn}{CNN}{Convolutional Neural Network}
\newacronym{map}{mAP}{mean average precision}
\newacronym{hps-net}{HPS-Net}{Holistic Parking Slot Network}
\newacronym{vpsd}{VPSD}{Valeo Parking Slots Dataset}
\newacronym{v2x}{V2X}{Vehicle-to-Everything}
\newacronym{sgd}{SGD}{Stochastic Gradient Descent}

\title{\LARGE \bf
Holistic Parking Slot Detection with Polygon-Shaped Representations
}

\author{Lihao Wang$^{1}$, 
        Antonyo Musabini$^{2}$, 
        Christel Leonet$^{2}$, 
        Rachid Benmokhtar$^{2}$, \\
        Amaury Breheret$^{3}$, 
        Chaima Yedes$^{2}$, 
        Fabian B\"urger$^{2}$, 
        Thomas Boulay$^{2}$ 
        and Xavier Perrotton$^{2}$\\
\thanks{$^{1}$Valeo Mobility Tech Center - Driving SoftWare and systems (DSW) - San Mateo - USA
        {\tt\small name.surname@valeo.com}}%
\thanks{$^{2}$Driving SoftWare and systems (DSW), Créteil - France
        }%
\thanks{$^{3}$Mines ParisTech - Center for Robotics, Paris - France}
}

\begin{document}

\maketitle
\thispagestyle{empty}
\pagestyle{empty}

%%%%%%%%%%%%%%%%%%%%%%%%%%%%%%%%%%%%%%%%%%%%%%%%%%%%%%%%%%%%%%%%%%%%%%%%%%%%%%%%
\begin{abstract}
Current parking slot detection in \gls{adas} primarily relies on ultrasonic sensors. This method has several limitations such as the need to scan the entire parking slot before detecting it, the incapacity of detecting multiple slots in a row, and the difficulty of classifying them. Due to the complex visual environment, vehicles are equipped with surround view camera systems to detect vacant parking slots. Previous research works in this field mostly use image-domain models to solve the problem. These two-stage approaches separate the 2D detection and 3D pose estimation steps using camera calibration. In this paper, we propose one-step \gls{hps-net}, a tailor-made adaptation of the \gls{yolo}v4 algorithm. This camera-based approach directly outputs the four vertex coordinates of the parking slot in topview domain, instead of a bounding box in raw camera images. Several visible points and shapes can be proposed from different angles. A novel regression loss function named polygon-corner \gls{giou} for polygon vertex position optimization is also proposed to manage the slot orientation and to distinguish the entrance line. Experiments show that \gls{hps-net} can detect various vacant parking slots with a F1-score of \bf{0.92} on our internal \gls{vpsd} and \bf{0.99} on the public dataset PS2.0. It provides a satisfying generalization and robustness in various parking scenarios, such as indoor (F1: 0.86) or paved ground (F1: 0.91). Moreover, it achieves a real-time detection speed of 17 FPS on Nvidia Drive AGX Xavier. A demo video can be found at \url{https://streamable.com/75j7sj}.

%(\gls{vpsd} has the longest covered range $\pm$12.5m compared to all known parking datasets)

% This method has several limitations such as the need for physical objects delimiting the slots (i.e. the line markings are not considered), the need to drive by and scan the entire parking slot before detecting it and finally the difficulty estimating the exact orientation of the slot. Nowadays, surround-view cameras have become low-cost and high-performance sensors that are installed around modern vehicles. Currently used for camera streams only, these cameras can be used to detect parking slots.

\end{abstract}

%%%%%%%%%%%%%%%%%%%%%%%%%%%%%%%%%%%%%%%%%%%%%%%%%%%%%%%%%%%%%%%%%%%%%%%%%%%%%%%%
\section{INTRODUCTION}\label{section:Introduction}

% Pr 1:  Why this is important? Numbers.
Parking is known to be one of the most stressful tasks for drivers. This is due to two major user pain points: finding a vacant parking slot and then executing the maneuver.

Over the past few decades, automotive companies have been developing various parking related \gls{adas} to address these issues, which are able to detect certain available slots and autonomously execute the parking maneuver. However, they failed to be accepted by end-users: 72\%~of drivers don't trust active parking-assist systems~\cite{AAA2015}. This result is due to the absence of efficient feedback on sensor data, a high degree of initial distrust in case of failure of execution (especially during the first few uses~\cite{Tenhundfeld2020}), the low speed of slot detection and automated maneuvers execution~\cite{Musabini2021}. As a result, the usability of current parking \gls{adas} remains weak.

% Pr 2: What is the problem ?
A number of parking \gls{adas} such as BMW's~Parking~Assistant~\cite{BMWUX}, Mercedes-Benz's~Parktronic~Active~Parking~Assist~\cite{PARKTRONIC}, Tesla~Model~X's~Park~Assist~\cite{ModelXManual}, Valeo's~Park4U~\cite{ValeoPark4u}, Bosch's~Parking~Aid~\cite{BoschParkingAid}, use ultrasonic sensors to locate available parking slots. However, a major limitation of these systems is that they can only detect parking slots after passing by them, as they need to scan the empty space between two parked vehicles to identify a vacant slot. Additionally, these systems rely on the presence of other vehicles parked while scanning, which means that they are not able to detect slots in empty parking areas or in front of successively empty parking slots. %In other words, in an empty parking area, they are not able to detect slots.

To address the limitations of ultrasonic sensors, recent studies (\cite{Wang2014, Zhang2018, Huang2019, Wu2020, Li2020}) resort to methods based on computer vision and deep learning. By generating topview images from surround view cameras and utilizing the RGB information of parking markings, these approaches are capable of detecting a wider range of use cases. However, most of them focus on individual parking markings and employ bottom-up or top-down approaches to obtain the final predictions, limiting the generalization capability of these methods across all parking types.

% Pr 3: What we propose ?
This work aims to overcome these limitations of the current systems by detecting vacant independent parking slots of all types, even in an empty parking lot. Our main contributions are:
\begin{itemize}
\item We propose a new polygon-shaped representation for camera-only parking slots detection that is able to cover all slot shapes (parallel, perpendicular, diagonal). This novel representation can efficiently model both fully visible and occluded slots, allowing for end-to-end learning.
\item We introduce a novel regression loss function named polygon-corner \gls{giou} for polygon corner position optimization. The proposed loss function is an approximation of the classical \gls{iou} loss yet is computationally efficient and enables parking entrance line prediction.
\item We train and test our approach \gls{hps-net} on diverse parking cases (normal, paving, indoor) with a wide distance range of 25m × 25m around the ego-vehicle. To the best of our knowledge, this coverage is the largest among all known parking detection methods, offering more choices for end-users.
% We use a new parking dataset called \gls{vpsd}. Each image of \gls{vpsd} covers an area of 25m × 25m around the ego-vehicle, which has the largest range compared against other known parking datasets. Furthermore, it includes all slot shapes (parallel, perpendicular, diagonal) in diverse user cases (normal, paving, indoor). 
\end{itemize}

% Feedback provided on detected parking slots ahead is expected to build trust on the end-users side and also to accelerate the execution of parking \glspl{adas}. We detect free parking slots in a range of 12.5~meters around the vehicle, even if they are partially occluded. The four surround view cameras of the vehicle are projected to a \gls{bev} image. This image is used as the input of a neural network, which is trained to locate free parallel, perpendicular and diagonal parking slots with a tailor-made parking slot representation. Four visible corners of each parking slots are located on the image, by distinguishing the entrance line. Finally, the outcome of the neural network is projected to the 3D world, in order to obtain the coordinates of the parking slot in the metric coordinate system.

Section~\ref{section:Related Work} details the state-of-the-art parking slot detection methods; section~\ref{section:Methodology} explains the proposed parking slot representation, the generation of the topview image, our custom dataset, the neural network algorithm details and the target platform which executes the model in a real vehicle. Then, section~\ref{section:Results} shows the obtained parking detection results, and finally section~\ref{section:Conclusions} presents the conclusions.

\section{RELATED WORK} \label{section:Related Work}

In general, parking space detection methods can be classified into four main categories~\cite{Ma2021}. These are:

\begin{itemize}
\item Free-space-based (\cite{BMWUX, PARKTRONIC, ValeoPark4u}): This group of methods scans the empty area of a parking slot with a distance measurement sensor such as ultrasonic sensors, light emitting sensors, 3D scanners, lidars, radars etc.
\item Parking-marking-based (\cite{Huang2019, Wu2020, Li2020}): This second group, with a camera, detects straight lines and corners before reconstructing present parking slots.
\item User-interface-based (\cite{HoGiJung2010}): These methods are more like semi-automatic detection systems, because they require an initial input from the user to specify a seed point for the target location.
\item Infrastructure-based: They rely on sensor-equipped infrastructures to guide the vehicle to an empty parking slot (e.g. Valeo's automated Valet Parking~\cite{ValeoValet}).
\end{itemize}

Free space based methods are not able to resolve the two main identified problems, as detailed in Section~\ref{section:Introduction}. User interface and infrastructure based methods might be available on only very high-end vehicles as they rely on very specific \gls{v2x} and human-machine communication methods. The only remaining method is using vehicle's vison based sensors (i.e. cameras) to detect parking slots. Therefore, the need is eliminated for additional sensors for the majority of brand new vehicles.

Most traditional computer vision-based methods detect parking markings from topview images. These detections can be for instance parking lines detected in Radon space~\cite{Wang2014}, parking corner detected with Harris corner detector~\cite{Suhr2014}, probabilistic reconstruction on detected edges and lines with Hough Transform~\cite{Hamada2015}. However, as with all traditional computer vision methods, these approaches are very sensitive to condition variation such as light, type and quality of markings.

\begin{figure}[thpb]
  \centering
  \subcaptionbox{PS2.0 \label{fig:existing_datasets:ps20}}{\includegraphics[height=.45\linewidth]{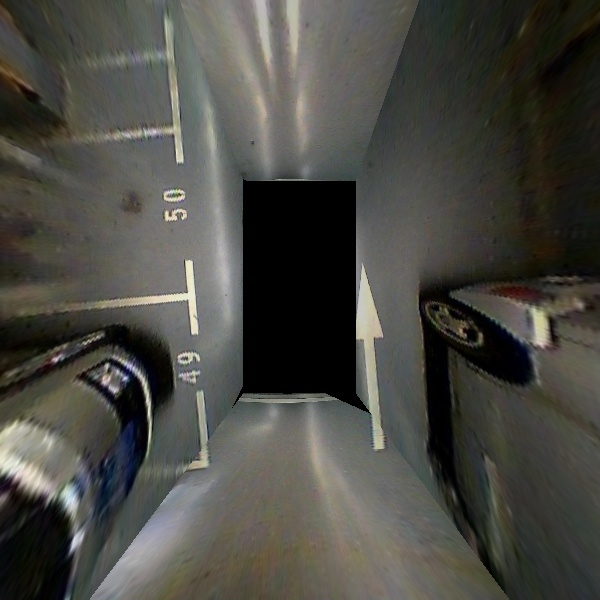}}\hspace{3em}%
  \subcaptionbox{PIL-Park SNU
  \label{fig:existing_datasets:snu}}{\includegraphics[height=.45\linewidth]{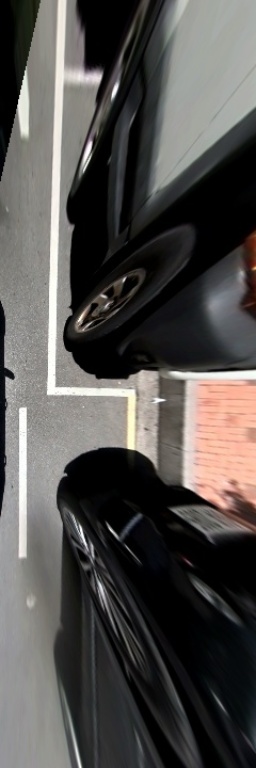}}
%\caption{{\bfseries Examples from publicly available datasets.} PS2.0 is reconstructed from full \gls{bev} of a projection range of 10m × 10m. SNU is reconstructed from half \gls{bev} of a projection range of 14.4m × 4.8m.}
\caption{{\bfseries Examples from publicly available datasets.} (a) PS2.0 with range of $\pm$10m; (b) PIL-Park/SNU with range of ($\pm$7.2m × 4.8m)}
\label{fig:existing_datasets}
\end{figure}

Recently, deep learning-based methods (\cite{Zhang2018, Huang2019, Wu2020, Li2020, Do2020}) also detect parking slots from topview space. PS2.0~\cite{Zhang2018} and PIL-Park/SNU~\cite{Do2020} are the most popular public topview-based parking slots datasets. PS2.0 includes T-shaped and L-shaped marking points (see Fig.~\ref{fig:existing_datasets:ps20}). Due to its low range 10m × 10m, in many conditions only the two entrance corners of the parking lot are visible, or only one vacant parking per image exists.
PIL-Park contains half topview images covering 14.4m × 4.8m of range (i.e. $\pm$7.2m × 4.8m) (see Fig.~\ref{fig:existing_datasets:snu}).

PS2.0~\cite{Zhang2018} has been used for many works such as DeepPS~\cite{Zhang2018}, which detects parking slots in two successive stages. According to their implementation, a first stage based on \gls{yolo}v2 detected two opposite parking corners as a rectangle and a second stage of another custom \gls{cnn} classified these corners as parking slots. However, this rectangular description is sensitive to the slots orientation. DMPR-PS~\cite{Huang2019} added the ability to estimate the orientation of parking with directional marking points, PSDet~\cite{Wu2020} proposed a circular template for marking points and Li et al.~\cite{Li2020} proposed a directional entrance line extractor to overcome the sensitivity to the direction changes. These methods performed well for only T and L shaped parking corners. VPS-Net~\cite{Li2020_VPS} was the first work able to detect and classify vacant as well as occupied parking slots. A first stage of \gls{yolo}v3-based detector and post-processing to pair the two detected opposite parking corners is used to locate parking slots and a custom \gls{cnn} as a second stage classified the occupancy of that parking. However this method can not perform well when the ego-vehicle is inside a parking slot and its orientation estimation is inaccurate due to the predefined orientations for diagonal slots.

%The cited past work used a publicly available parking image dataset called 
%PS2.0~\cite{Zhang2018} and PIL-Park/SNU~\cite{Do2020} are the most popular public topview-based parking slots datasets. PS2.0 includes T and L shaped corners (see Fig.~\ref{fig:existing_datasets:ps20}). Due to its low range -/+ 5m, in many conditions only the two entrance corners of the parking lot are visible, or {\color{red}only} one vacant parking per image {\color{red}exists}.
%PIL-Park contains half topview images covering 14.4m×4.8m of range {\color{red} (-/+7.2x4.8)} (see Fig.~\ref{fig:existing_datasets:snu}).

Do et al.~\cite{Do2020}, by using the PIL-Park/SNU dataset, proposed a two stage method: (i) parking context recognizer to detect the slot's orientation; (ii) \gls{yolo}v3 locates slots as bounding boxes. These bounding boxes are rotated with the orientation detected during the first stage. The outcome is the coordinates of the four vertices of a parking slot. This method performs well in most cases, but it assumes that adjacent parking slots have the same orientation and type, which is not necessarily true.

More recent works combine diverse techniques to first detect slots roughly and then refine detections (e.g. global and local information extractor~\cite{Suhr2021} or region proposal network and slot detection / classification networks \cite{Bui2021}). It is also worth mentioning that some detection algorithms using directly the raw image are investigated~\cite{Xu2020}, even with the distinction of the entrance line~\cite{patentNvidia}.

None of the existing datasets provide sufficient coverage zone to allow those algorithms to achieve a human-like detection capability. To address this limitation, we have collected and annotated a new parking dataset covering an area of 25m × 25m (i.e. $\pm$12.5m with ego-car in the center of the topview image) around the ego-vehicle (see Section~\ref{section:Methodology:Dataset}). As far as we know, this is the largest coverage for \gls{adas} parking assistant.

% $25m\times 25m$

\section{METHODOLOGY} \label{section:Methodology}

\subsection{Polygon Representation} \label{section:Methodology:PolygonRepresentation}

\begin{figure}[thpb]
  \centering
  \subcaptionbox{Perpendicular\label{fig:perpendicular}}{\includegraphics[width=.3\linewidth]{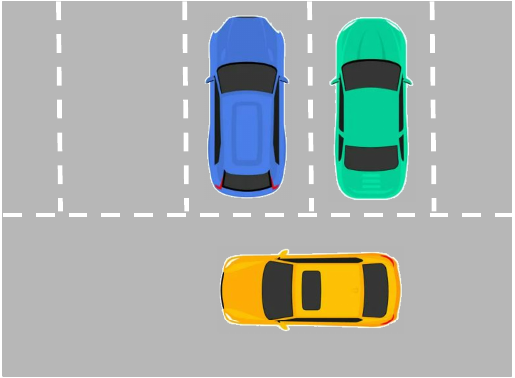}}\hspace{1em}%
  \subcaptionbox{Parallel\label{fig:parallel}}{\includegraphics[width=.3\linewidth]{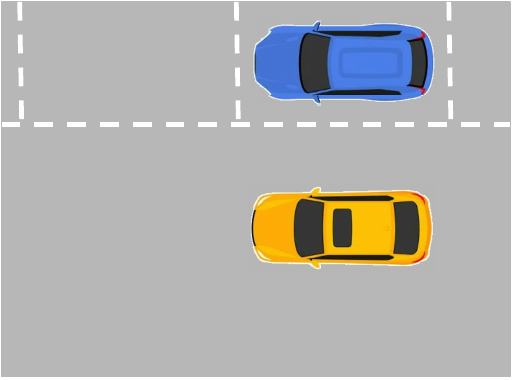}}\hspace{1em}%
  \subcaptionbox{Diagonal\label{fig:fishbone}}{\includegraphics[width=.3\linewidth]{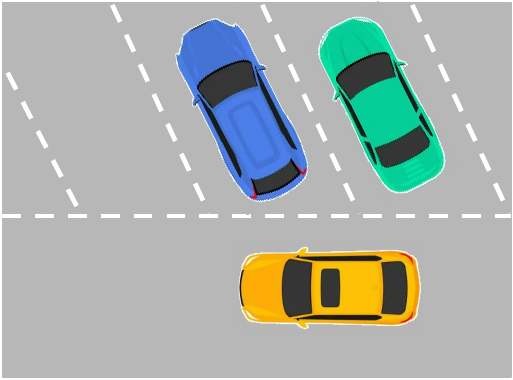}}
\caption{Different types of parking slots}
\label{fig:parking_slot_types}
\end{figure}

Based on their shape, parking slots can be generally classified into three categories: perpendicular, parallel and diagonal (also known as fishbone or slanted slot), like illustrated in Fig. ~\ref{fig:parking_slot_types}. Each of them contains four key-points (i.e. the four corners) which indicate delimitation of the parkable area. These four key-points can be hence considered as the necessary and sufficient elements to define one parking slot.

\begin{figure}[thpb]
  \centering
  \subcaptionbox{Global\label{fig:global_representation}}{\includegraphics[width=.45\linewidth]{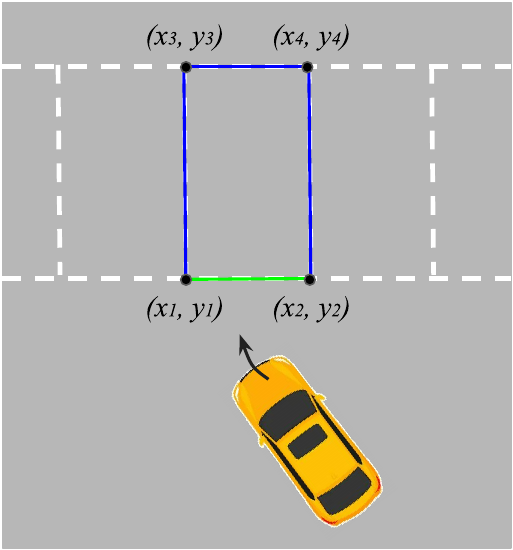}}\hspace{1em}%
  \subcaptionbox{Relative\label{fig:relative_representation}}{\includegraphics[width=.45\linewidth]{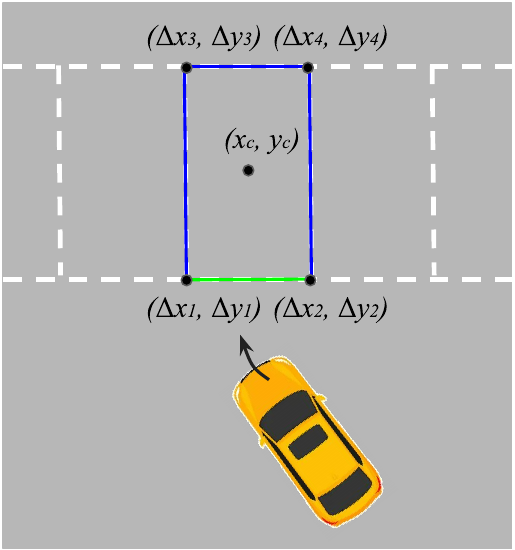}}
%\caption{{\bfseries Polygon representation.} Green line signifies the entrance line.}
\caption{{\bfseries Polygon representation}. The entrance line is in green.}
%Green line signifies the entrance line.}
\label{fig:global_polygon_representation}
\end{figure}

%(Fig. ~\ref{fig:global_representation})
Unlike other topview-based approaches \cite{Zhang2018, Huang2019, Wu2020, Li2020}, which work only for certain parking types, our work aims at finding one general model which covers all shapes. To this end, a four-point polygon (also known as quadrangle or quadrilateral) model is considered to represent each parking slot, as shown in Fig. ~\ref{fig:global_polygon_representation}. It is defined by a vector of four coordinates as follows:

\begin{equation}
  \label{eq:global_polygon_vector}
  \begin{aligned}[b]
    \{({x}_{1}, {y}_{1}), ({x}_{2}, {y}_{2}), ({x}_{3}, {y}_{3}), ({x}_{4}, {y}_{4})\},
    \end{aligned}
\end{equation}

\noindent where $(x_{1}, y_{1})$, $(x_{2}, y_{2})$, $(x_{3}, y_{3})$, $(x_{4}, y_{4})$ signify the entrance-left, entrance-right, ending-left, ending-right points, respectively. The entrance line consists of two points, entrance-left and entrance-right, marking the side from which the car should be parked into. In some rare cases, a slot may contain more than one line that can be entered, then the entrance line can be defined as any one of them. And the ending line is defined as the opposite side of the entrance line. It is worth noting that the order of the four corners should be respected because the entrance and ending lines information is critical during the path planning phase.

In practice, to facilitate the learning process of \gls{dnn}, we use the following relative representation:
\begin{equation}
  \label{eq:relative_polygon_vector}
  \begin{aligned}[b]
    \{({x}_{c}, {y}_{c}), ({{\Delta}x}_{1}, {{\Delta}y}_{1}), ({{\Delta}x}_{2}, {{\Delta}y}_{2}), ({{\Delta}x}_{3}, {{\Delta}y}_{3}), ({{\Delta}x}_{4}, {{\Delta}y}_{4})\}
    \end{aligned}
\end{equation}
% where $(x_{c}, y_{c})$ signifies the center point of the parking slot. % like:
%\begin{equation}
%  \label{eq:polygon_center}
%  \begin{aligned}[b]
%    {x}_{c} =\frac{1}{4} \ {\sum_ {i=1}^{4} {x}_{i}}\hspace{3em}%
%    {y}_{c} =\frac{1}{4} \ {\sum_ {i=1}^{4} {y}_{i}}
%    \end{aligned}
%\end{equation}

\noindent where $(x_{c}, y_{c})$ signifies the center point of the parking slot and $({\Delta}x_{1}, {\Delta}y_{1})$, $({\Delta}x_{2}, {\Delta}y_{2})$, $({\Delta}x_{3}, {\Delta}y_{3})$, $({\Delta}x_{4}, {\Delta}y_{4})$ are the coordinate offsets of entrance-left, entrance-right, ending-left, ending-right points from the center point, respectively. 

\subsection{Topview representation}\label{susection:Methodology:TopView}

\gls{ipm} is widely employed in self-driving applications (e.g. lane detection~\cite{6957662}). It assumes a plane world (height $z=0$). Then, with the calibration parameters, it maps all pixels from a given viewpoint onto this flat plane through homography projection. 

% The \gls{bev} created with \gls{ipm} provides a correct result of the ground, while all objects on the ground will be deformed (as observed with the white vehicle in Fig. ~\ref{fig:topview}). In our case there are four images (front, rear, left, right - see Fig. ~\ref{fig:SurroundViewImages}) with which we construct four \glspl{ipm} and finally merge them into one. Empirically we have chosen to keep the left and right images which cover 180° on both sides, then complete the rest with the front and rear ones (see Fig. ~\ref{fig:topview}).

\begin{figure}[thpb]
    \centering
    \subfloat[\centering Front fisheye image]{\includegraphics[width=.445\linewidth]{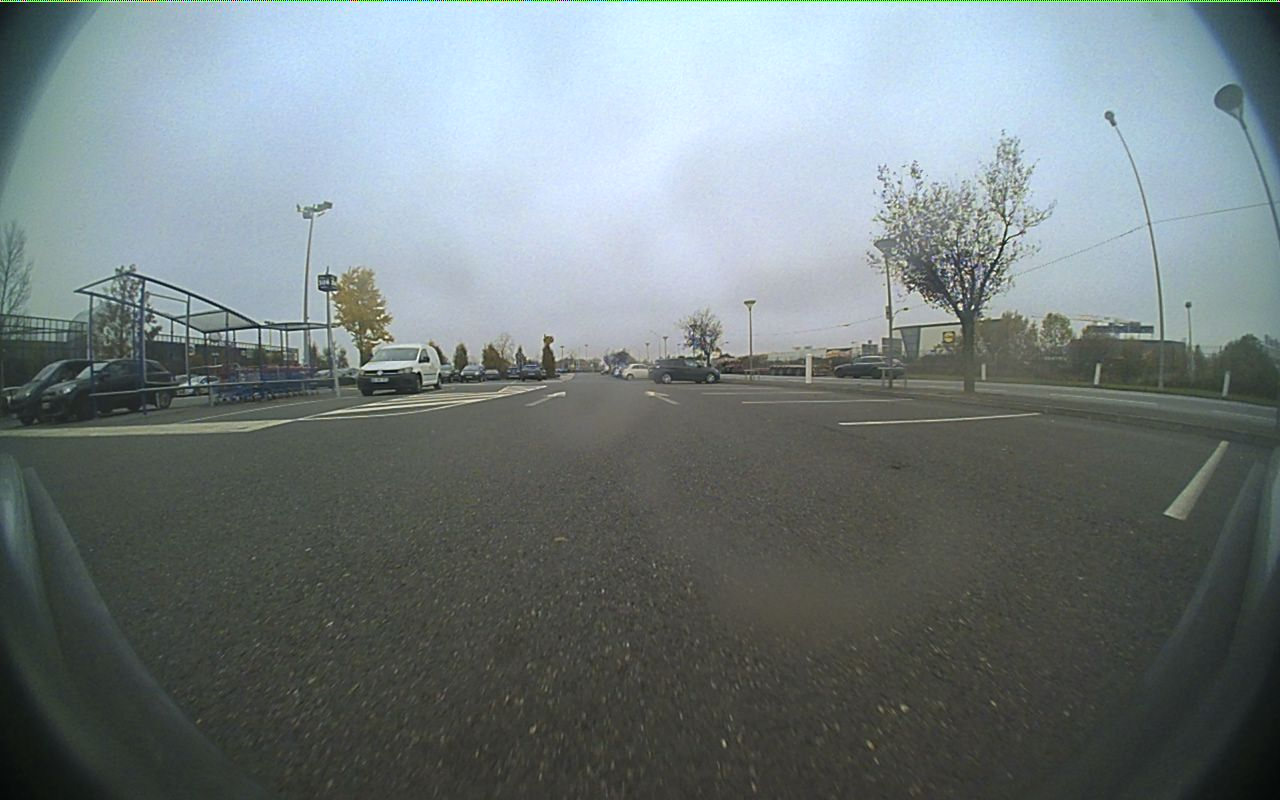}\label{fig:front} }
    \qquad
    \subfloat[\centering Rear fisheye image]{\includegraphics[width=.445\linewidth]{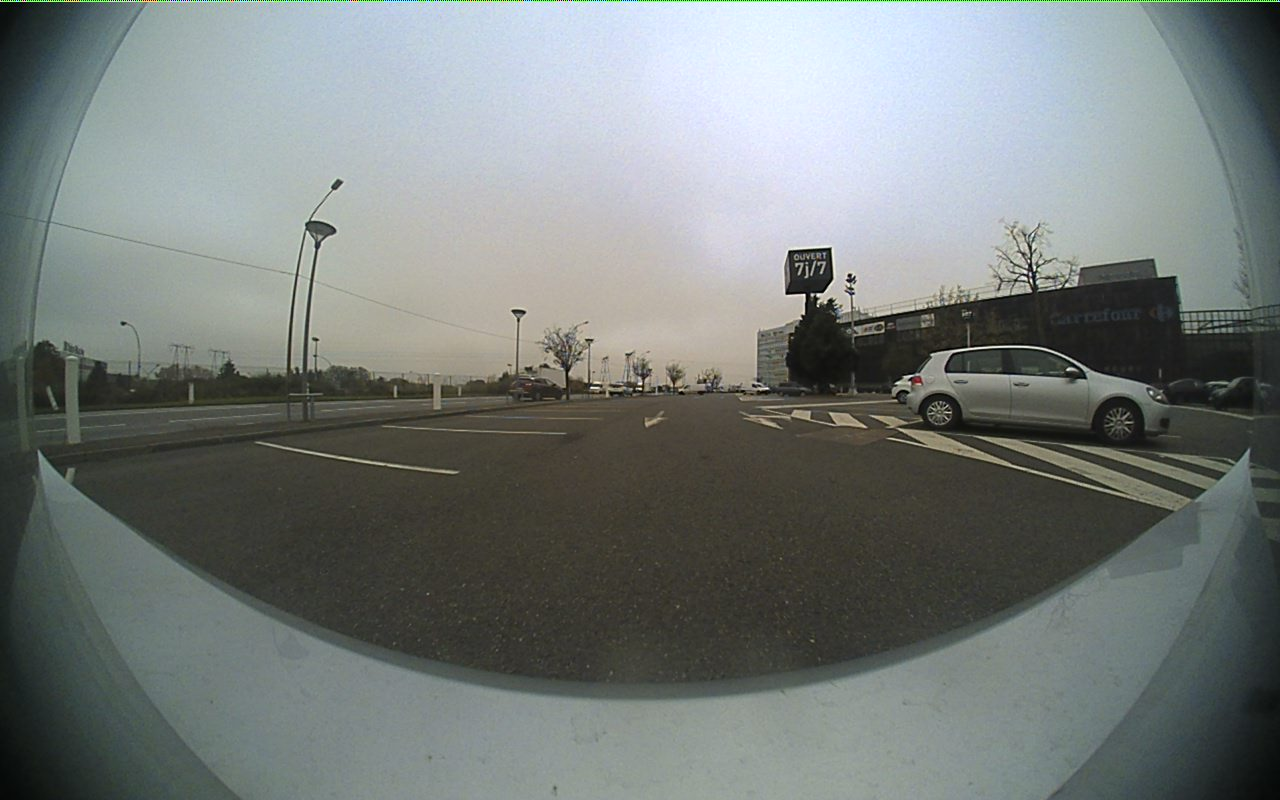}\label{fig:rear} }
    \qquad
    \subfloat[\centering Left fisheye image]{\includegraphics[width=.445\linewidth]{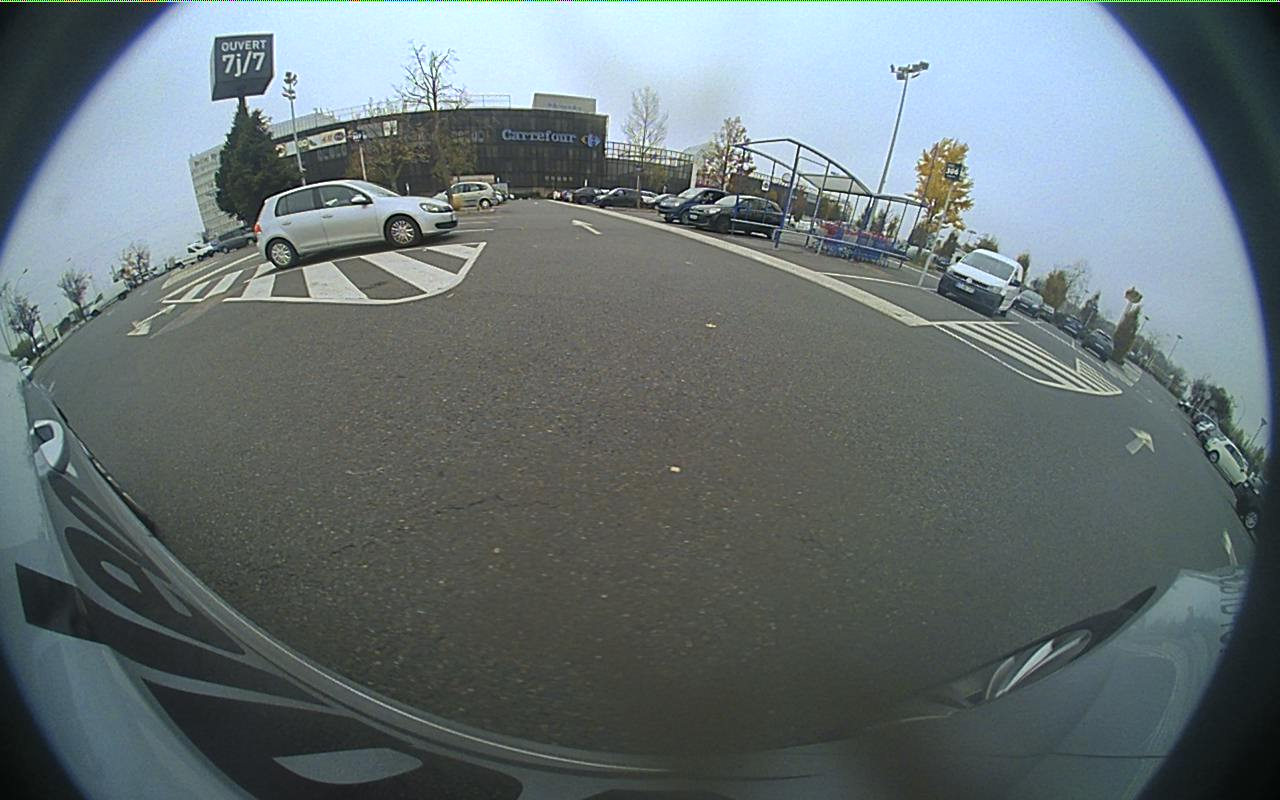}\label{fig:Left} }
    \qquad    
    \subfloat[\centering Right fisheye image]{\includegraphics[width=.445\linewidth]{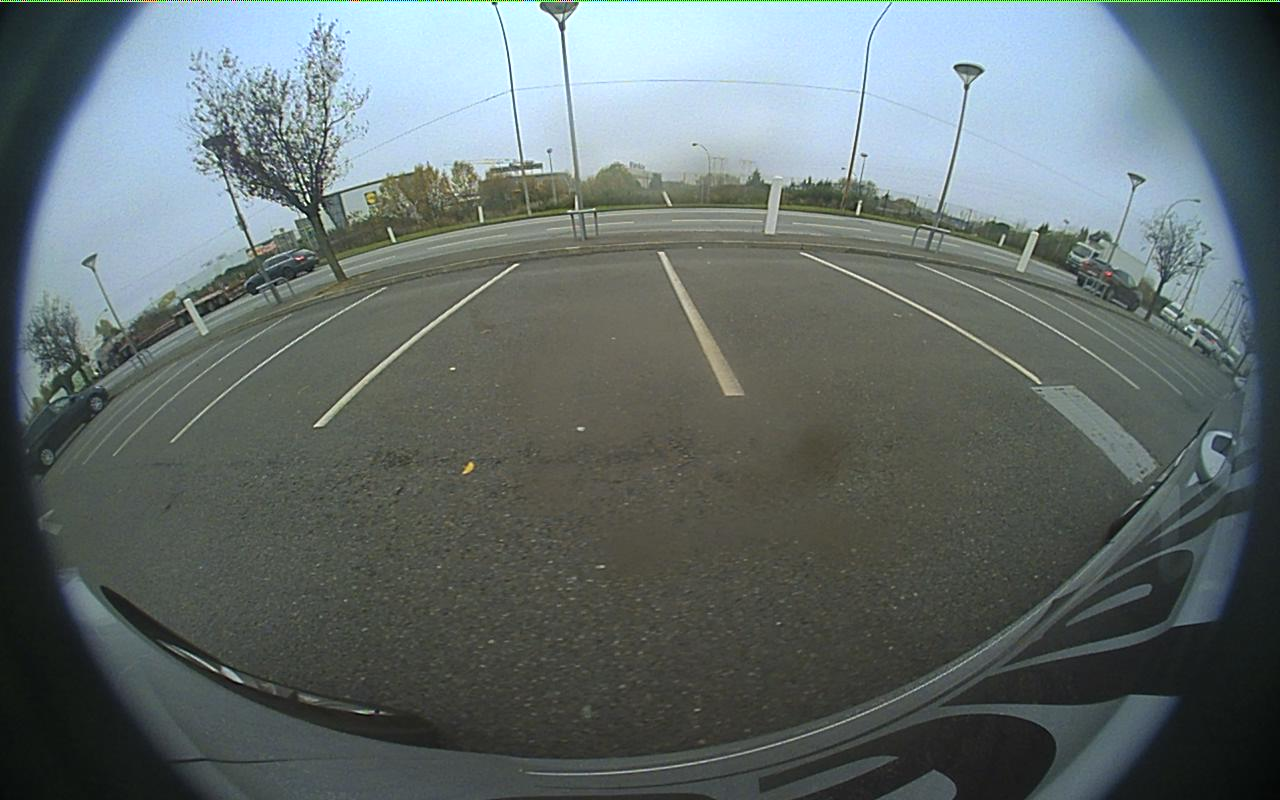}\label{fig:right} }
    \caption{Surround view images from four fisheye cameras of a vehicle}    
    \label{fig:SurroundViewImages}    
\end{figure}

\begin{figure}[thpb]
    \centering
    \subfloat[\centering Created topview image from Fig.~\ref{fig:SurroundViewImages}]{\includegraphics[width=.445\linewidth]{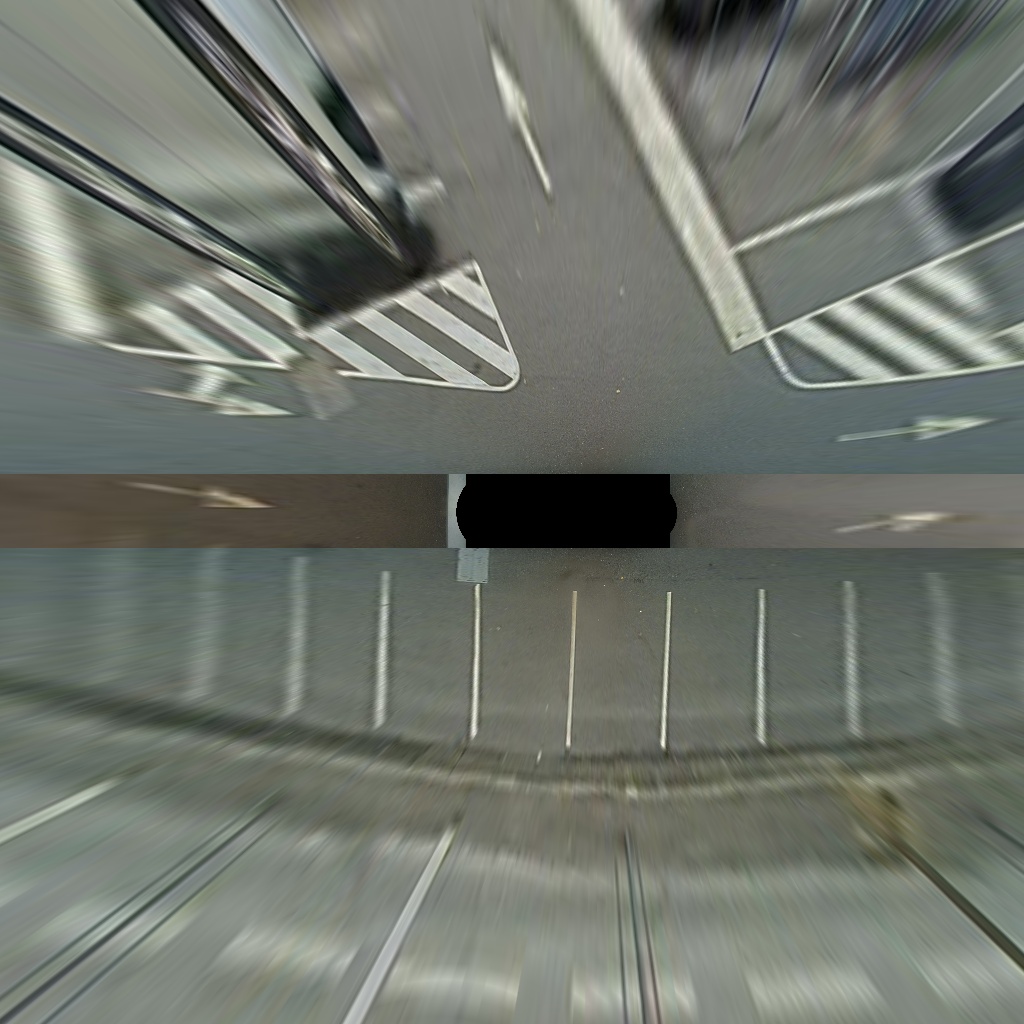}\label{fig:topview} }
    \qquad
    \subfloat[\centering Detected parking slots]{\includegraphics[width=.445\linewidth]{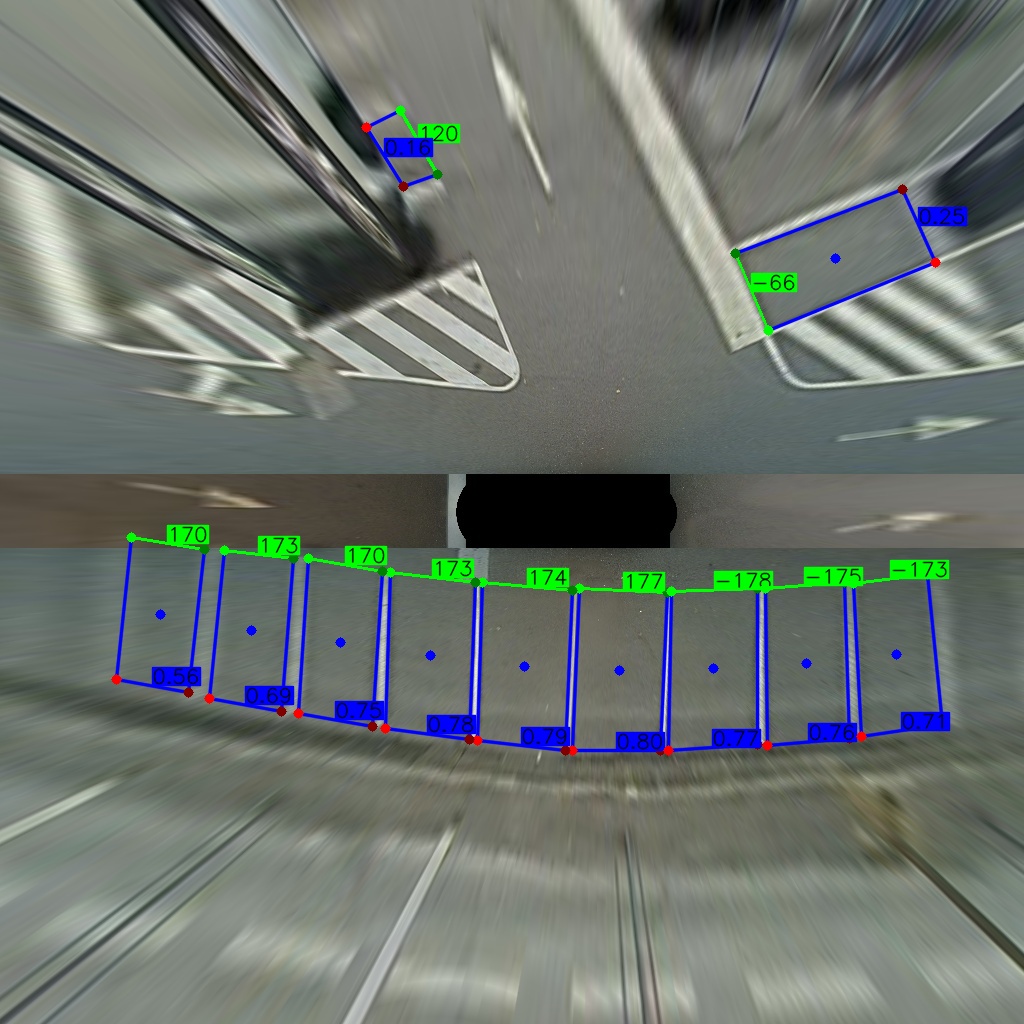}\label{fig:detections} }
    \caption{Topview image and and the overlay of parking slot detections}
    \label{fig:TopViewDetections}    
\end{figure}

The created topview provides a correct projection of the ground, while all objects are deformed (see  Fig. ~\ref{fig:topview}). In our work, four surround view cameras are mounted on the vehicle (front, rear, left, right - see Fig. ~\ref{fig:SurroundViewImages}). IPM is applied on each camera to get four individual topview images which are then merged into one global topview image (see  Fig. ~\ref{fig:topview}). %Empirically we have chosen to keep the left and right images which cover 180° on both sides, then complete the rest with the front and rear ones (see Fig. ~\ref{fig:topview}).

\subsection{Datasets}\label{section:Methodology:Dataset}

In order to evaluate the performance, we conduct experiments on the following two datasets.

\subsubsection{Self-annotated VPSD dataset}
\gls{vpsd} is constructed to train and evaluate our proposed solution. Surround view fisheye images are collected from typical indoor and outdoor parking sites using six different demo cars. The fisheye image resolution is 1280 × 800. The \gls{ipm} topview image resolution is 640 × 640, which covers a 25m × 25m flat surface. This means that each pixel corresponds to a length of 3.9cm on the physical ground.
Polygon-shaped parking slots have been annotated and verified from the topview images by at least two human annotators. The \gls{vpsd} contains 11,227 topview images in the training set and 2,737 topview images in the testing set, resulting in a total of 104,330 polygon-shaped boxes. %This dataset is not yet publicly available.%

%The annotated \gls{vpsd} contains 11,227 \gls{bev} images in the training set and 2,737 \gls{bev} images in the testing set, resulting in a total of 104,330 polygon-shaped boxes. This dataset is not yet publicly available.

We define the \textit{normal} parking as the one that is located outside, on asphalt ground. Two other specific use cases are also considered during data collection: these are slots located inside buildings (\textit{indoor}) and slots on brick-paved ground (\textit{paving}). Table~\ref{table:VPSD_by_use_case} and table~\ref{table:VPSD_by_slot_type} list the numbers of images per use case and numbers of polygon-shaped boxes per slot type, respectively. Fig. ~\ref{fig:parking_types} showcases examples from different use cases. %and slots which have at least one kerb boundary (\textit{kerbs})

\begin{table}[h]
    \caption{Image numbers by use case in \gls{vpsd} dataset}
    % v5
    \begin{center}
    \begin{tabular}{cc}
        \hline
        \textbf{\textit{Subset}}  &  \textbf{\textit{Number of images}} \\
        \hline
        Normal & 9,757     \\ \hline  
        Indoor & 2,097     \\ \hline       
        Paving & 2,110    \\ \hline
        Total & 13,964    \\ \hline
    \end{tabular}
    \end{center}
    \label{table:VPSD_by_use_case}
\end{table}

\begin{table}[h]
    \caption{Box numbers by parking slot type in \gls{vpsd} dataset}
    % v5
    \begin{center}
    \begin{tabular}{cc}
        \hline
        \textbf{\textit{Parking slot type}}  &  \textbf{\textit{Number of boxes}} \\
        \hline
        perpendicular  & 100,928  \\ \hline
        parallel & 1,576  \\ \hline
        diagonal  & 1,826  \\ \hline
        Total & 104,330 \\ \hline
    \end{tabular}
    \end{center}
    \label{table:VPSD_by_slot_type}
\end{table}

% Various locations and conditions (indoors/outdoors, sunny, night, snow, etc.)
% Multiple vehicles / cameras

\subsubsection{PS2.0 dataset}
In order to provide a fair comparison against the state-of-the art approaches, we also evaluated our algorithm on PS2.0~\cite{Zhang2018} dataset.%, which is the only publicly available dataset on full \gls{bev} parking images.

\subsection{Neural Network Training}\label{section:Methodology:Training}

Section~\ref{section:Related Work} referred to many previous works, which used different versions of \gls{yolo} to locate parking slots in images. One of the main reasons for this common choice is the speed of execution, due to the fact that it performs in only one stage. Hence, we also have chosen \gls{yolo}v4~\cite{yolov4} as the starting point to develop our custom polygon-based variant algorithm to detect parking slots. It is worth noting, however, that our solution could be integrated into any other object detection DNN.

\begin{figure}[thpb]
  \centering
  \subcaptionbox{Axis-aligned bounding box\label{fig:bbox}}{\includegraphics[width=.3\linewidth]{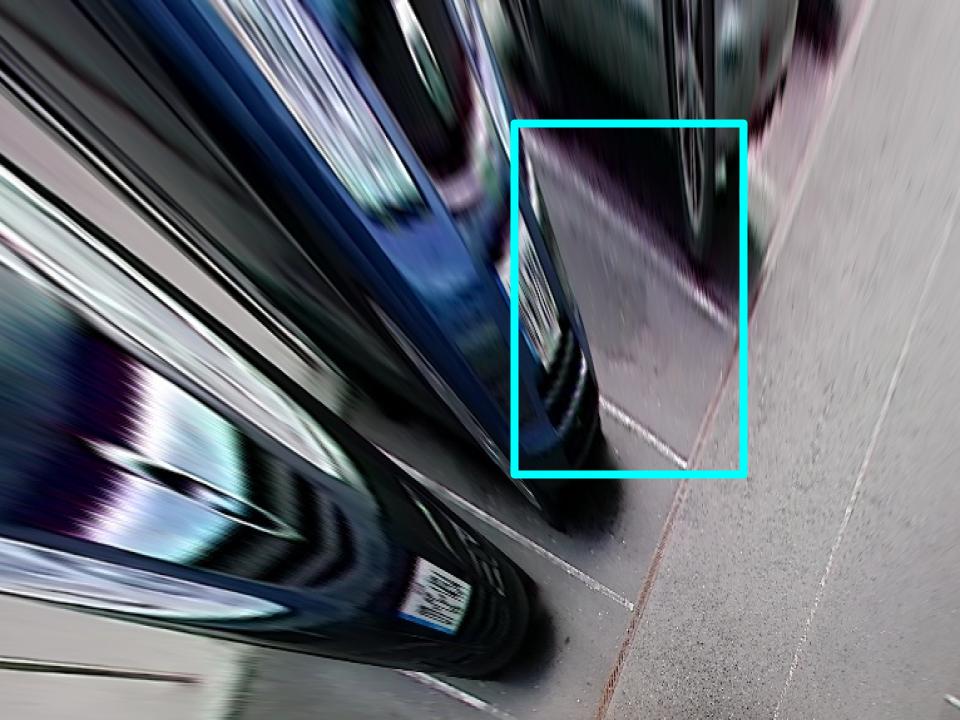}}\hspace{1em}%
  \subcaptionbox{Oriented bounding box\label{fig:oriented_box}}{\includegraphics[width=.3\linewidth]{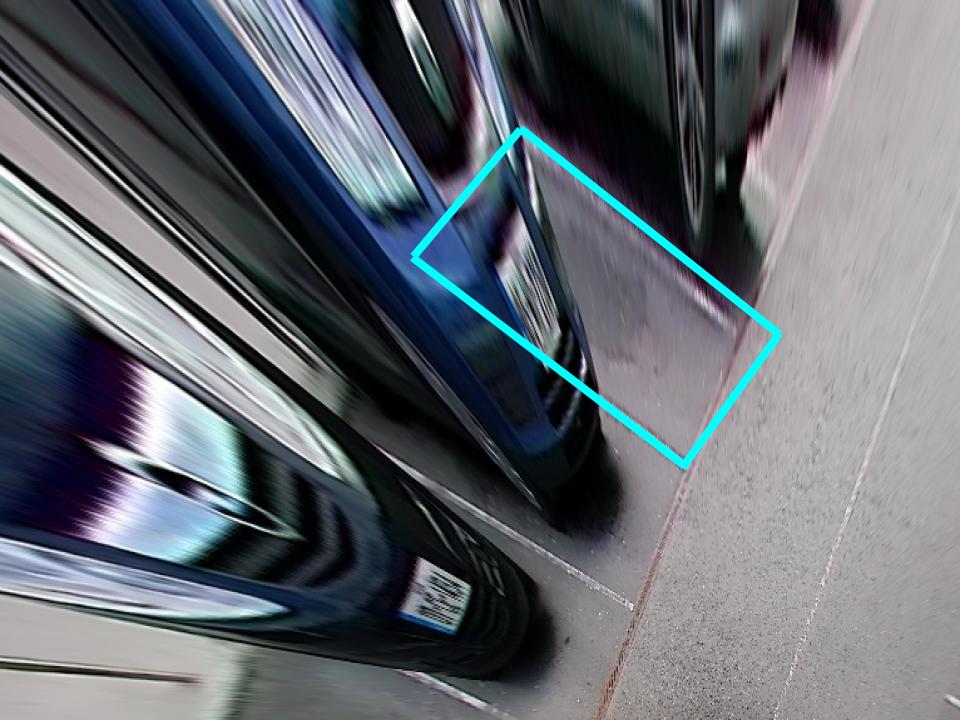}}\hspace{1em}%
  \subcaptionbox{Polygon (ours)\label{fig:polygon}}{\includegraphics[width=.3\linewidth]{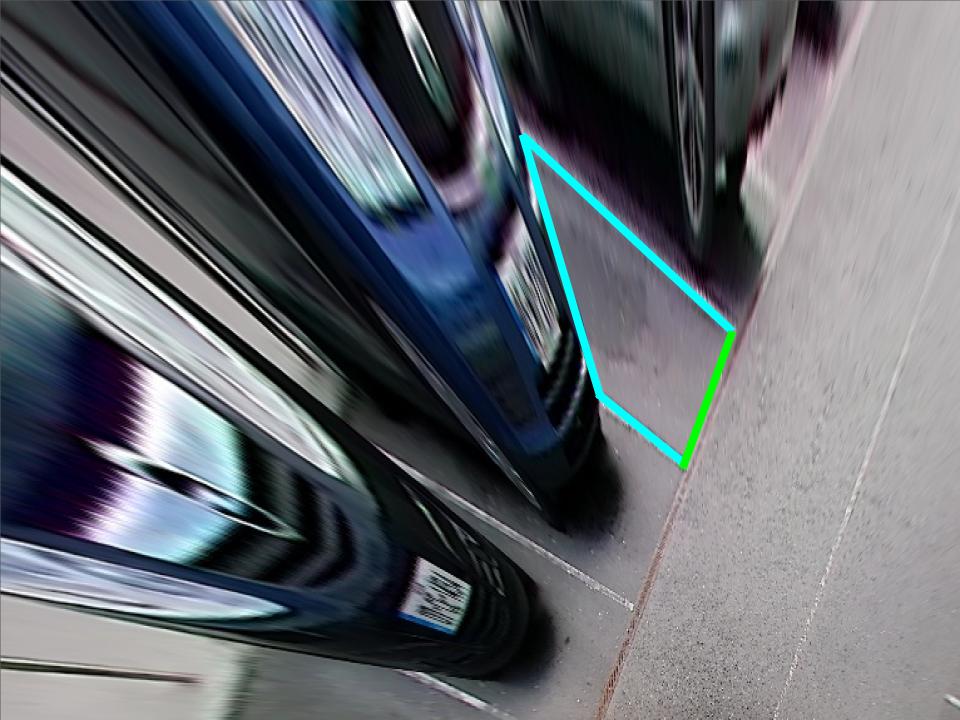}}
\caption{{\bfseries Different types of prediction boxes}}
\label{fig:box_types}
\end{figure}

Common 2D detection algorithms (including \gls{yolo}v4) predict an object's location with axis-aligned or rotated rectangles, called bounding boxes. Instead of using a bounding box, we resort to a four-point polygon to represent a parking slot, as described in Section~\ref{section:Methodology:PolygonRepresentation}. The main reason is that for diagonal slots or partially occluded slots, the visible corners form a four-point polygon of any shape. Hence, bounding box representation, even rotated, is not suitable. On the contrary, our proposed polygon representation can still precisely fit the visible area in occlusion cases, as shown in Fig.~\ref{fig:box_types}.

\begin{figure}[thpb]
    \centering
    {\includegraphics[width=.5\linewidth]{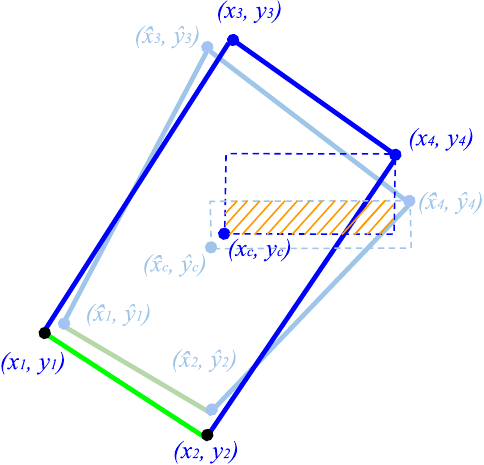}}
    \caption{{\bfseries Polygon-corner \gls{giou}}. Ground truth is in dark colors and the prediction is in light colors. For simplicity, only one corner is picked to show the related bounding boxes, where the intersection part is in orange.}
    \label{fig:polygon_iou}
\end{figure}

The loss function is important for \gls{dnn} models' convergence  during training and generalization capability. \gls{yolo}v4 uses several variants of \gls{iou} loss such as \gls{giou}~\cite{giou}, ~\gls{diou} and ~\gls{ciou}~\cite{diou}. \gls{giou} takes into account the proximity between the prediction and the ground truth, showing a constant improvement over standard \gls{iou}. However, these \gls{iou}-based loss functions only work with axis-aligned bounding boxes, and a straightforward thought is to implement a similar polygon \gls{iou} loss function for our case. Unfortunately, this solution has two main limitations: firstly, its implementation is non-trivial due to the irregular shape of arbitrary polygons, and secondly, none of these \gls{iou}-based loss functions consider the order of the box corners, which is critical for predicting the entrance line. To overcome these challenges, we propose a novel loss function named polygon-corner \gls{giou}, which provides an efficient approximation of polygon \gls{giou}. Concretely, polygon-corner \gls{giou} considers a four-point polygon as a group of four bounding boxes, each one is formed by the polygon's center and one corner (see Fig.~\ref{fig:polygon_iou}). 
%The \gls{giou}~\cite{giou} computes the minimum convex area which covers the detection and the actual object's location, which is useful to score bad predictions as well. Hence, we developed an approximation of \gls{giou}, applicable to our polygon shaped representations.

Thus the final \gls{giou} loss is the mean value of the four bounding box \gls{giou} losses:
\begin{equation}
  \label{eq:polygon_giou}
  \begin{aligned}[b]
    % \mathcal{L}_{GIoU\ polygon} = \frac{1}{4} * {\sum_ {bbox=1}^{4} GIoU_{bbox}}
    \mathcal{L}_{GIoU} = \frac{1}{4} {\sum_ {bbox=1}^{4} GIoU_{bbox}}
    \end{aligned}
\end{equation}

\noindent according to~\cite{giou}, each $GIoU_{bbox}$ is defined as:

\begin{equation*}
GIoU_{bbox} = IoU_{bbox} - \frac{\left| C \backslash \left( A \cup B \right) \right|}{|C|}
\end{equation*}

\noindent where $A$ and $B$ represent the predicted and ground truth bounding boxes, respectively. $C$ is the smallest bounding box that completely encloses both $A$ and $B$. And $IoU_{bbox}$ is the standard \gls{iou} function:

\begin{equation*}
IoU_{bbox}= \frac{|A\cap B|}{|A\cup B|}
\end{equation*}

One main advantage of polygon-corner \gls{giou} is that it retains the computational efficiency of classical \gls{giou} hence will not slow down the training process. Using \gls{giou} rather than \gls{iou} can reflect if two shapes are in vicinity of each other or very far. Additionally, this loss function takes into account the order of each corner, ensuring that the predicted corners are arranged in the same sequence as the ground truth. This facilitates the extraction of the entrance line at a later stage. 

Inspired by ~\gls{diou}~\cite{diou}, we also introduced an auxiliary loss term in the regression loss function to make it more stable. This loss calculates the distances between the four predicted corners and the corresponding ground-truth corners:

%The~\gls{diou}~\cite{diou} computes the distance between the center point of the prediction and the ground truth to resolve the same problem. However the~\gls{diou} is not suited to our multi-shaped polygon representations.

\begin{equation}
  \label{eq:lpixel}
    % \mathcal{L}_{pixel} = \frac{1}{4} * {\sum_ {j=1}^{4} \sqrt{ (x_{pred}^{i,j} - x_{gt}^{i,j})^2 + (y_{pt}^{i,j} - y_{gt}^{i,j})^2 }}
    \mathcal{L}_{dist} = \frac{1}{4} \ {\sum_ {i=1}^{4} \sqrt{(x_{i} - \hat{x_{i}})^2 + (y_{i} - \hat{y_{i}})^2 }}
\end{equation}

\begin{table*}[!htb]%!htb
    % Single Cls - v5
    % error of 2.57 pixels in average -> 5cm
    \caption{\gls{hps-net} results on \gls{vpsd} test set}% \gls{giou} in this table actually refers to polygon-corner \gls{giou}.}
    % a test set of 19,939 parking slots from 2737 images (19,248 parallel, 384 perpendicular, 307 diagonal
    \begin{center}
    \begin{tabular}{ccccccccc}
        \hline
        \textbf{\textit{Activation}}  & \textbf{\textit{Loss function}}  & \textbf{\textit{Precision}} & \textbf{\textit{Recall}} & \textbf{\textit{F-1 Score}}  & \textbf{\textit{mAP@.5}} & \textbf{\textit{mAP@.5:.95}} \\ %& \textbf{\textit{M. Pixel Err.}}
        \hline
        % Mish & Corner dist & 0.824 & 0.829 & 0.826 & 0.743 & 0.401 \\ \hline
        Mish & Corner dist & 0.844 & 0.849 & 0.846 & 0.768 & 0.471 \\ \hline
        % Mish & GIoU & 0.870 & 0.908 & 0.888 & 0.888 & 0.681 \\ \hline
        Mish & GIoU & 0.887 & 0.925 & 0.906 & 0.909 & 0.772 \\ \hline
        % Mish & GIoU \& Corner dist & \textbf{0.897} & \textbf{0.906} & \textbf{0.901} & \textbf{0.886} & \textbf{0.679}  & \\ \hline    
        Mish & GIoU \& Corner dist & \textbf{0.915} & \textbf{0.924} & \textbf{0.919} & \textbf{0.907} & \textbf{0.770}  \\ \hline % & 2.57
        %ReLu & GIoU \& Corner dist & 0.875 & 0.917 & 0.896 & 0.894 & 0.687 & \\ \hline
        ReLU & GIoU \& Corner dist & 0.892 & 0.935 & 0.913 & 0.914 & 0.777 & \\ \hline        
    \end{tabular}
    \end{center}
    \label{table:Results}
\end{table*}
% Equation~\ref{eq:lpixel} depicts the pixel wise distance between four detected corners to their ground truth.

% \begin{equation}
%   \label{eq:ldistance}
%     \mathcal{L}_{cor.\ distance} = \overline{max(\num{1e-16}, min({L}_{pixel}^{i}, Thr.)) / Thr. }
% \end{equation}

% Equation~\ref{eq:ldistance} depicts the corner distance loss, which is the mean of pixel distance of four points, after clamping, threshold and scaling.

% The~\gls{bcewll} is also computed to detect the objectness of parking slots, as in the original~\gls{yolo}v4 implementation.

Finally, the polygon box regression loss is a weighted sum of the two elements described earlier: the polygon-corner~\gls{giou} and the auxiliary distance loss:

% \begin{align}
%     \mathcal{L}_{polygon} = &\omega_{giou} * (1-\mathcal{L}_{polygon\ GIoU}) + \nonumber\\
%                   &\quad  \omega_{objectness} * \gls{bcewll} + \nonumber\\
%                   &\quad  \omega_{corner\ distance} * \mathcal{L}_{corner\ distance}\label{eq:loss}
% \end{align}

\begin{equation}
    \mathcal{L}_{polygon} = \omega_{giou} (1-\mathcal{L}_{GIoU}) + \omega_{dist} \mathcal{L}_{dist}\label{eq:loss}
\end{equation}

With regards to the classification loss, we used the same one (i.e. binary cross-entropy) as in \gls{yolo}v4~\cite{yolov4}.

% It is worth to mention that, different from the original \gls{yolo}v4 implementation the CIoU (an \gls{iou} derivative which takes into consideration of the ratios of detected shapes) and the \gls{bcewll} for class distinction weren't used (due to the shapes of our polygons and the fact that we have a single class problem).

While training, topview images are randomly transformed using one or more of following data augmentation techniques: left-right flipping, upside-down flipping, rotation of a random angle between 0° and 25°, \gls{hsv} color-space adjustment. The random data augmentation constantly generates new unseen images to the network and prevents over-fitting. Polygon box regression loss weights $\omega_{giou}$ and $\omega_{dist}$ are set to 1 and 0.75, respectively. Training on \gls{vpsd} is performed by \gls{sgd} optimizer for 500 epochs with a learning rate of $10^{-2}$, momentum 0.9, weight decay $5 \times 10^{-4}$ and batch size 16. The training took 3 days with a Nvidia Tesla V100 graphics card.

\subsection{Embedded Platform}\label{section:Methodology:EmbeddedPlatform}

The ultimate goal of developing this algorithm is to integrate it into vehicles. Nvidia's Drive AGX Xavier is an embedded computer, which is adapted to this purpose. It delivers industry-leading performance and energy-efficient computing for the development and production of functionally safe AI-powered cars, trucks, robotaxis, and shuttles~\cite{drive-agx}.

For this end, our initial module, trained in Python, needs to be firstly accelerated with Nvidia's TensorRT module to take advantage of embedded hardware. Second, it has to be implemented in a compiled computer language (e.g. C++) and be executed on such embedded platforms. However, due to the compatibility issues with embedding the algorithm, some further modifications in the architecture of \gls{yolo}v4 had to be done. These are replacing all \textit{Mish}~\cite{mish} activation functions with standard \textit{ReLU} activation and replacing all the up-sampling layers with inverse convolutions. Section~\ref{section:Results} shows the effects of these modifications.

\section{RESULTS}\label{section:Results}

Fig. ~\ref{fig:detections} illustrates an example of our holistic parking slots prediction results, where each blue polygon represents one detected slot. The entrance line is highlighted in green and its angle with respect to the ego-vehicle is printed in the same color. A confidence value per slot is also displayed in blue.

Table~\ref{table:Results} shows the quantitative results of our detection method over a test set of 2,737 images, containing in total 19,939 parking slots (19,248 parallel, 384 perpendicular, 307 diagonal). Precision, recall, F-1 score and \gls{map} scores are calculated in different configurations. It is visible that the mandatory modifications needed for embedded systems (see Section~\ref{section:Methodology:EmbeddedPlatform}) resulted in a slight drop in precision and a slight gain in recall.

Ablation test for different parts of our loss function (see Table~\ref{table:Results}) is also conducted. When the polygon-corner \gls{giou} is deactivated during the training, the resulting model's precision and recall have dropped by {\bf{6\%}} and {\bf{7\%}} respectively. And deactivating the pixel-distance loss causes a precision drop by {\bf{3\%}}. Hence, further tests have been conducted under the full loss configuration.

Table~\ref{table:ResultsOtherTypes} details the detection metrics on test sets containing only specific use cases (\textit{indoor} and \textit{paving}). It is noticeable that the presence of pavings does not perturb the algorithm's accuracy, whereas the indoor use case results in a 14\% decrease in precision. Two possible reasons for this lower precision in indoor scenarios are the insufficient illumination in some indoor areas and light reflections on the ground that resemble parking lines.

\begin{figure*}[h]
    \centering
    %\subfloat[\centering Outdoor, asphalt]{\includegraphics[width=.2\linewidth]{img/dataset/395.jpg}\label{fig:395} }
    %\qquad
    %\subfloat[\centering Detected parking slots from \ref{fig:topview}]%{\includegraphics[width=.2\linewidth]{img/dataset_detections/395_detections.jpg}%\label{fig:395_detections} }
    \subfloat[\centering Outdoor, asphalt]{\includegraphics[width=.2\linewidth]{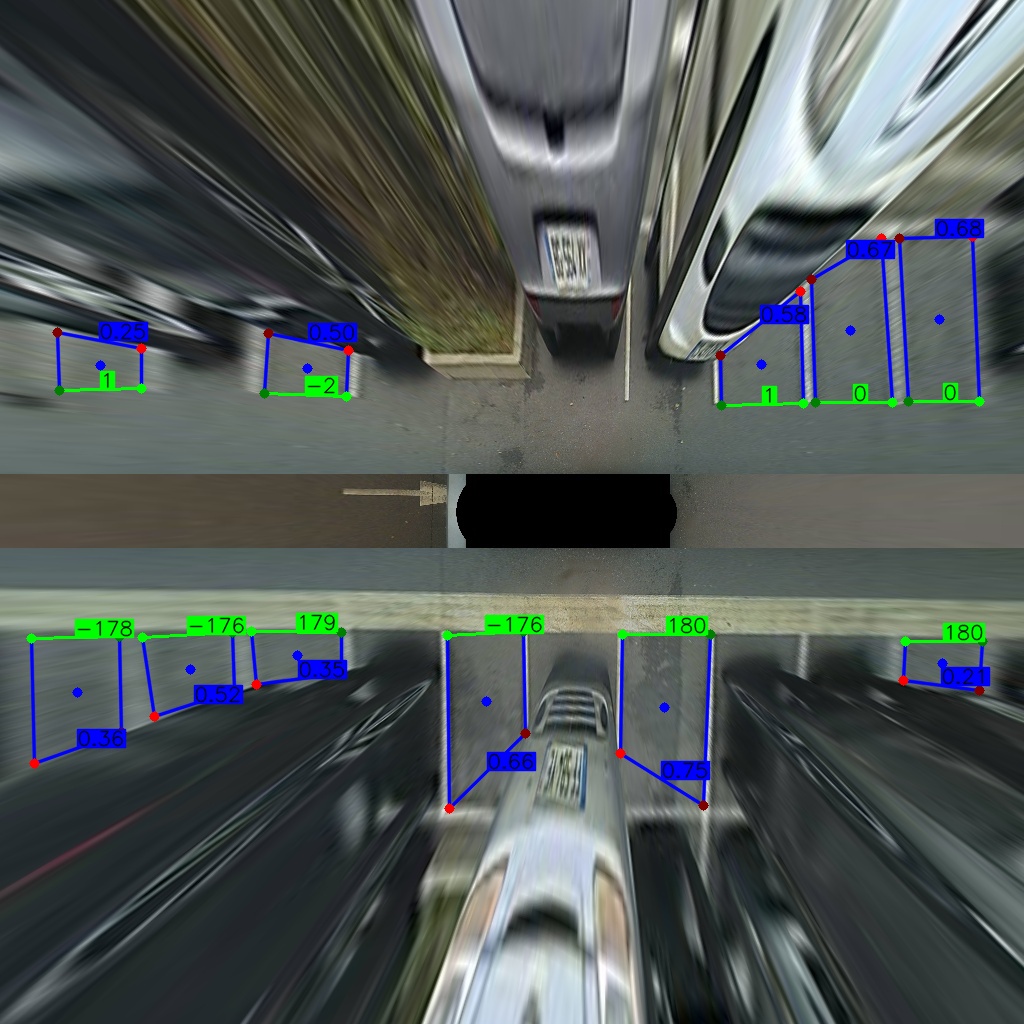}\label{fig:395_detections} }    
    \qquad
    %\subfloat[\centering Outdoor, asphalt]{\includegraphics[width=.2\linewidth]{img/dataset/road_1.jpg}\label{fig:road_1}}
    %\qquad
    \subfloat[\centering Outdoor, asphalt]{\includegraphics[width=.2\linewidth]{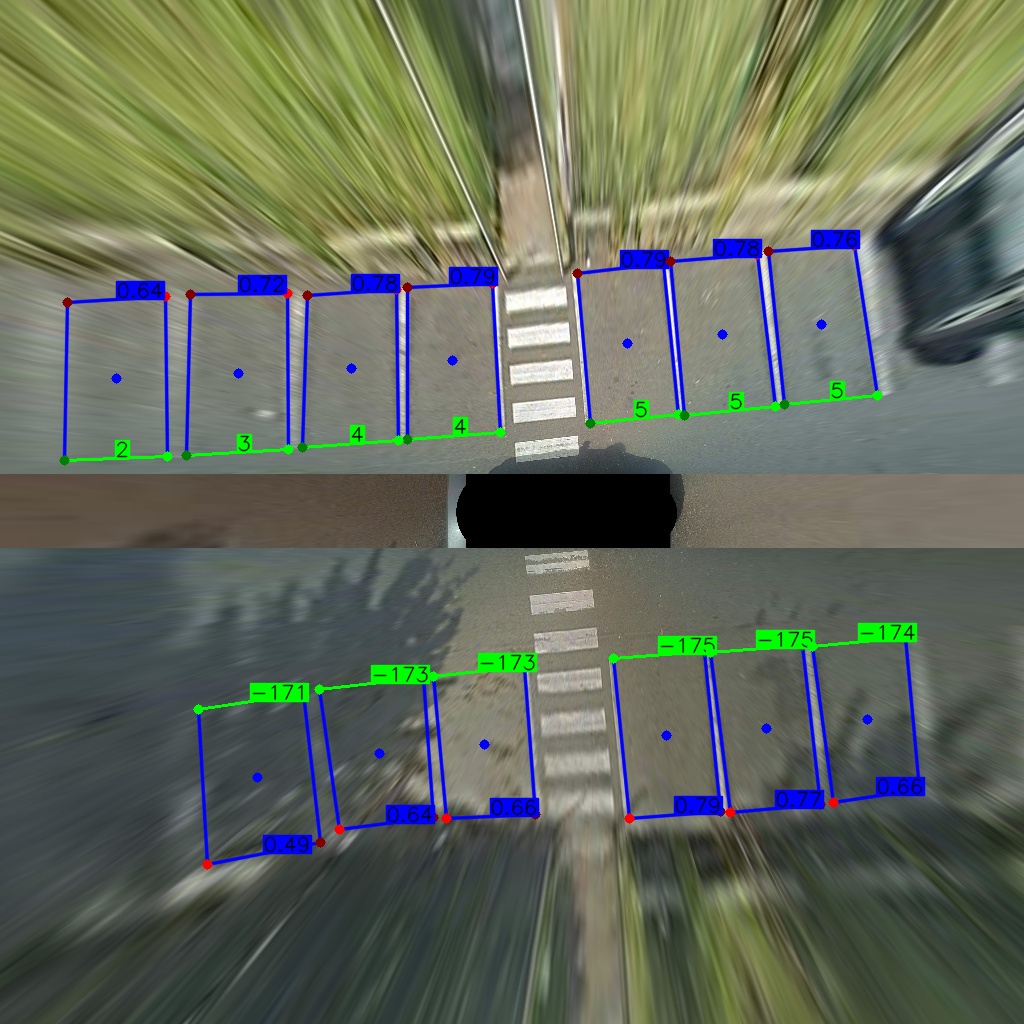}\label{fig:road_1_detections} }
    \qquad
    %\subfloat[\centering Outdoor, paving]{\includegraphics[width=.2\linewidth]{img/dataset/paving_3.jpg}\label{fig:paving_3} }
    %\qquad   
    \subfloat[\centering  Outdoor, paving]{\includegraphics[width=.2\linewidth]{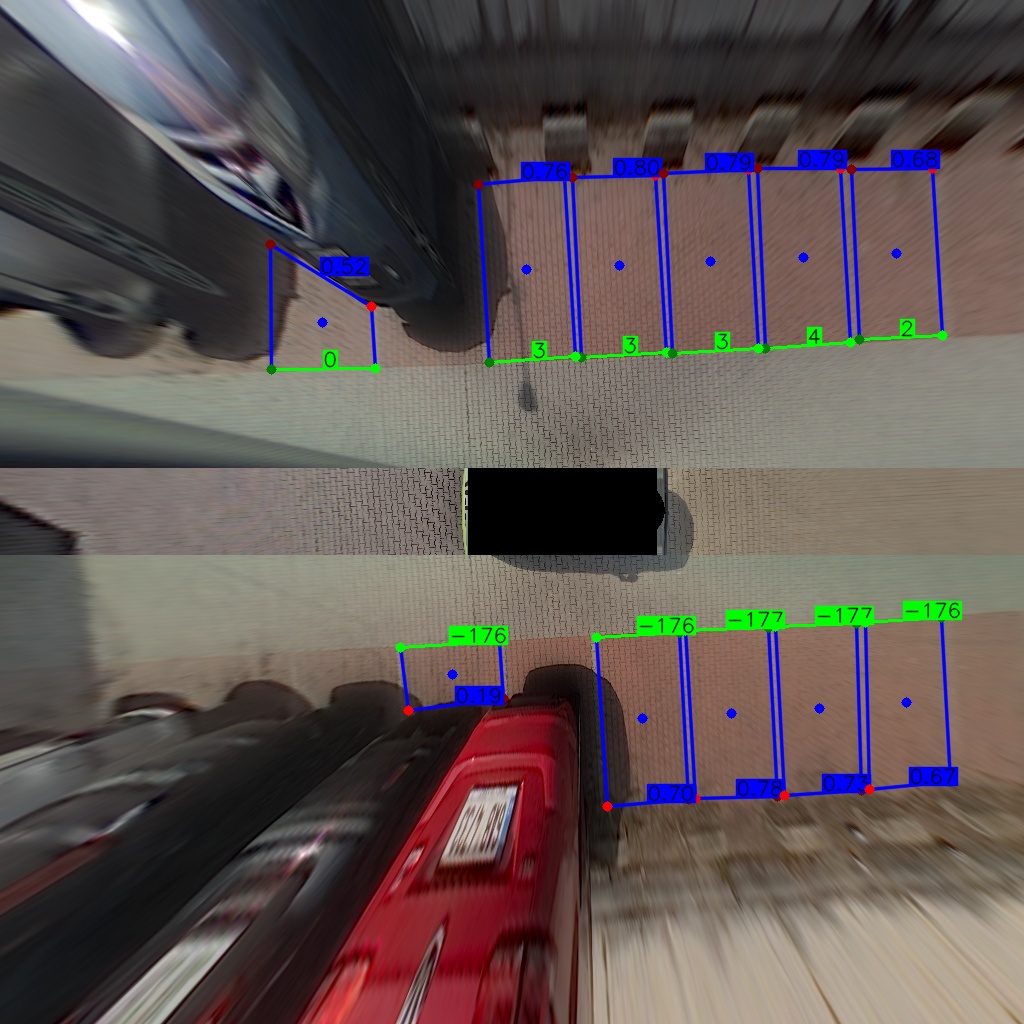}\label{fig:paving_3_detections} }
    \qquad
    %\subfloat[\centering Outdoor, paving]{\includegraphics[width=.2\linewidth]{img/dataset/paving_5.jpg}\label{fig:paving_5} }
    %\qquad   
    \subfloat[\centering  Outdoor, paving]{\includegraphics[width=.2\linewidth]{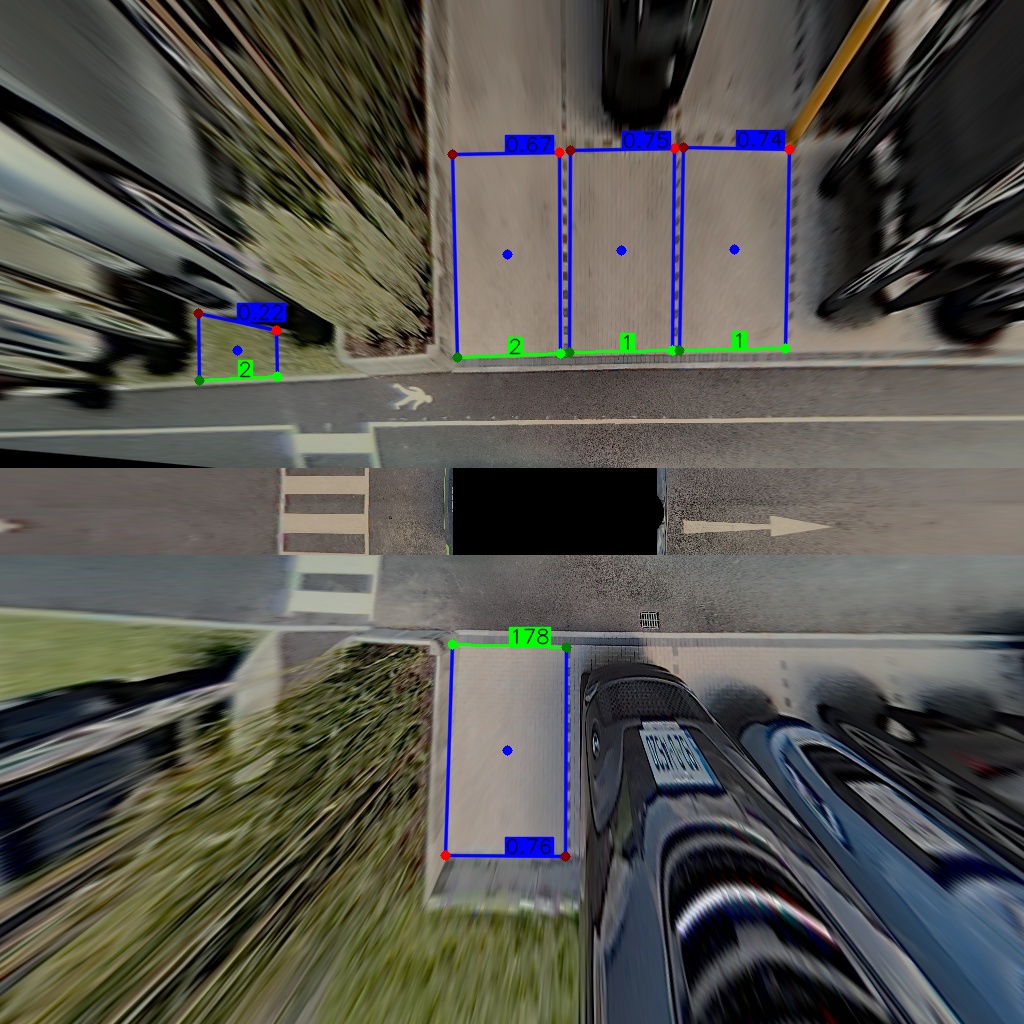}\label{fig:paving_5_detections} }
    \qquad
    %\subfloat[\centering Indoor]{\includegraphics[width=.2\linewidth]{img/dataset/indoor_1.jpg}\label{fig:indoor_1} }
    %\qquad    
    \subfloat[\centering Indoor]{\includegraphics[width=.2\linewidth]{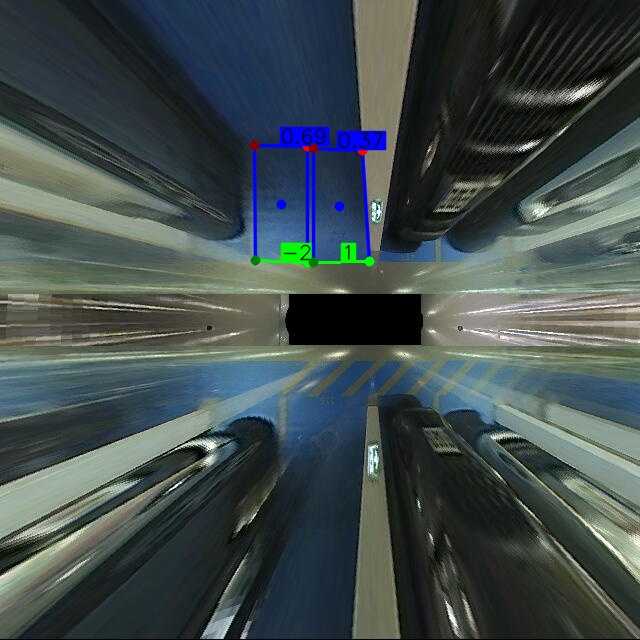}\label{fig:indoor_1_detections} }
    \qquad   
    %\subfloat[\centering Indoor]{\includegraphics[width=.2\linewidth]{img/dataset/indoor_2.jpg}\label{fig:indoor_2} }
    %\qquad    
    \subfloat[\centering Indoor]{\includegraphics[width=.2\linewidth]{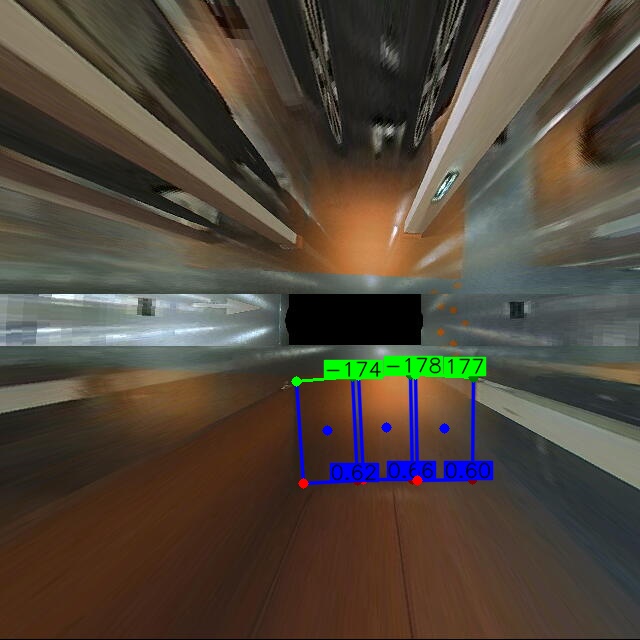}\label{fig:indoor_2_detections} }
    \qquad     
    %\subfloat[\centering Kerb (the right one)]{\includegraphics[width=.2\linewidth]{img/dataset/kerb_1.jpg}\label{fig:kerb_1} }
    %\qquad 
    \subfloat[\centering Kerb]{\includegraphics[width=.2\linewidth]{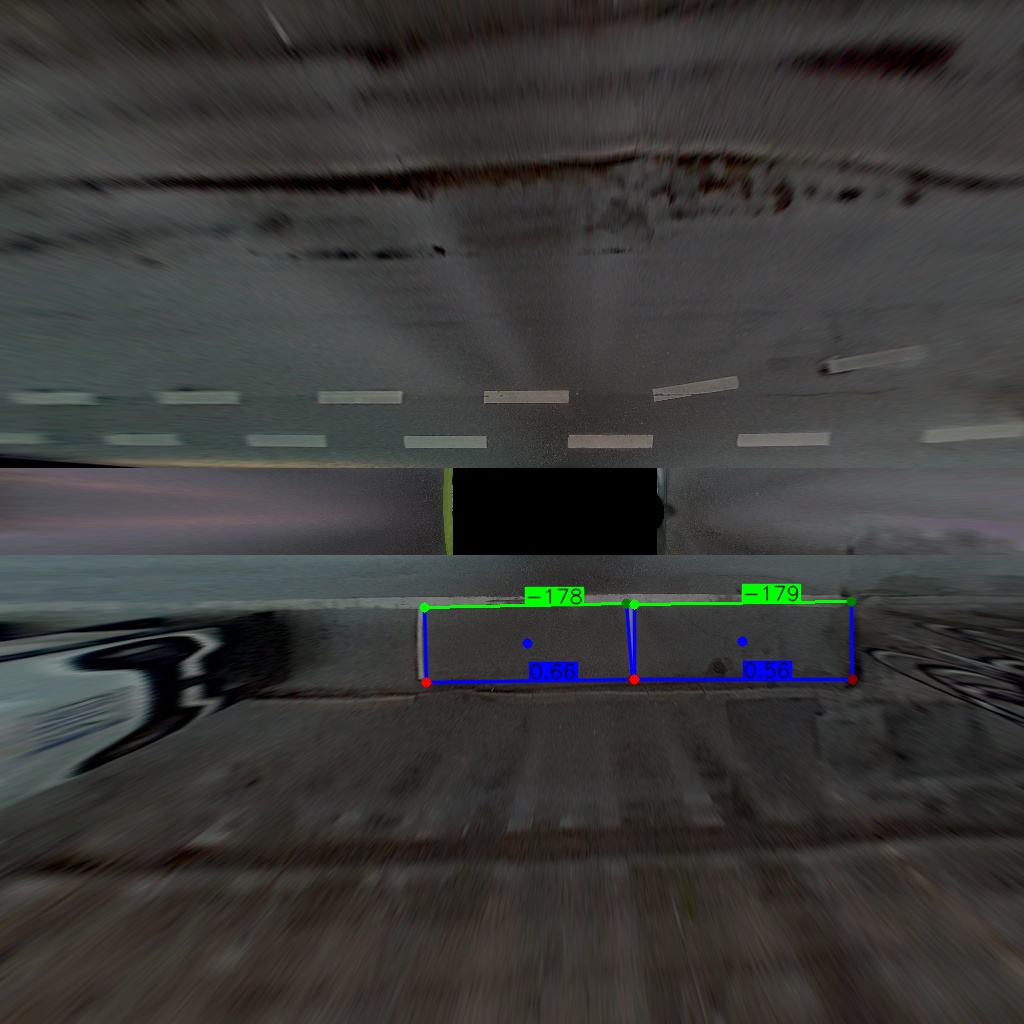}\label{fig:kerb_1_detections} }
    \qquad
    %\subfloat[\centering Multiple orientations]{\includegraphics[width=.2\linewidth]{img/dataset/277.jpg}\label{fig:277} }
    %\qquad 
    \subfloat[\centering Multiple orientations]{\includegraphics[width=.2\linewidth]{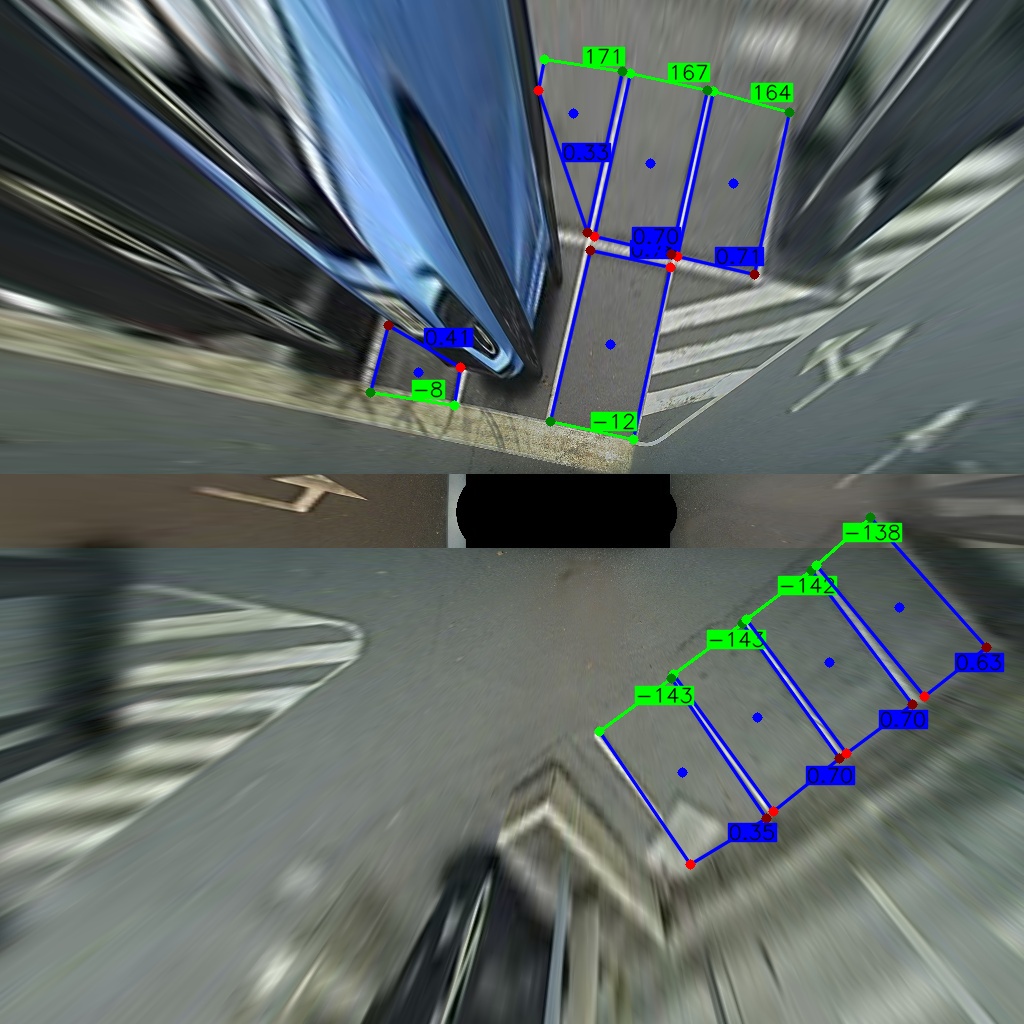}\label{fig:277_detections} }
    \qquad
    \caption{ Different types of parking slots in \gls{vpsd} with detection results.}
    \label{fig:parking_types}
\end{figure*}

\begin{figure*}[h]
    \centering
    \subfloat{\includegraphics[width=.2\linewidth]{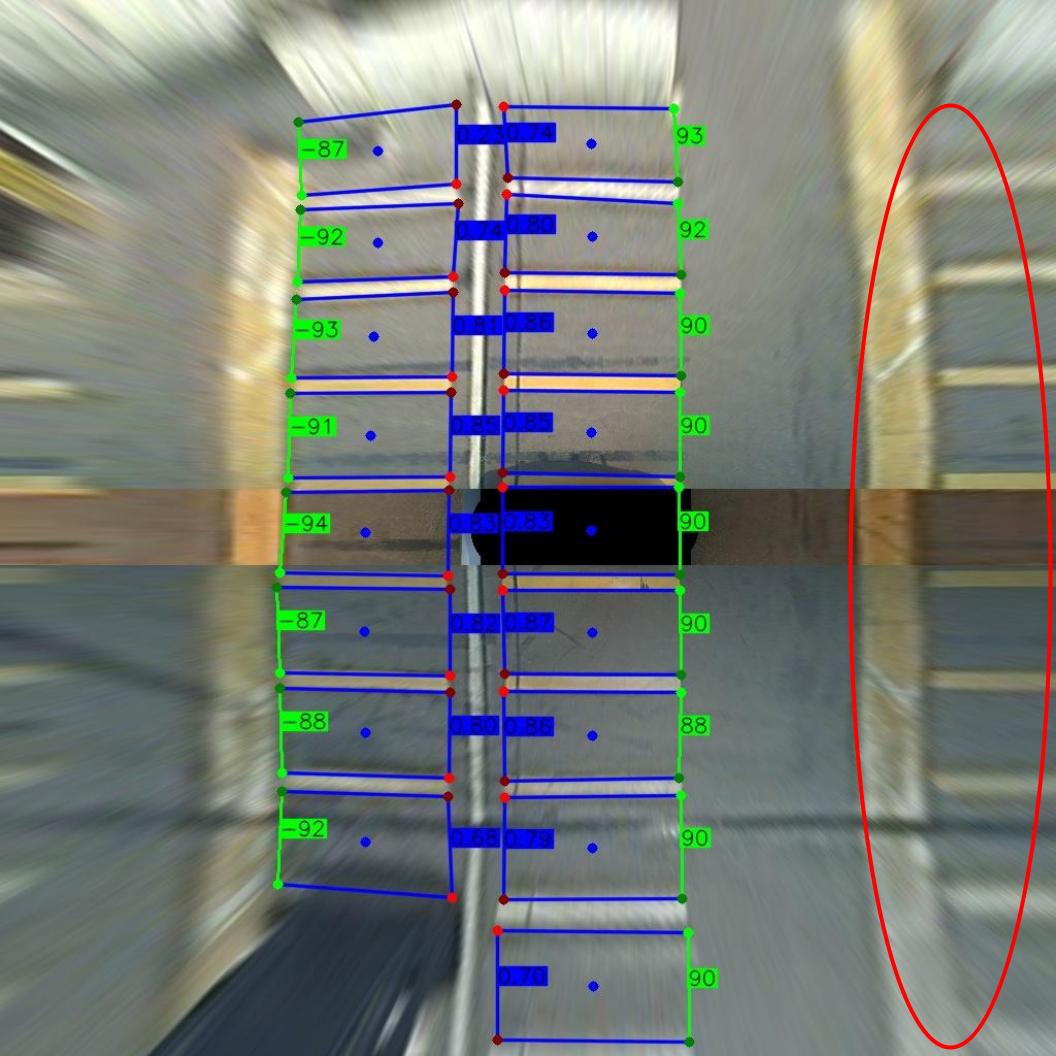}\label{fig:FN1} }    
    \qquad
    \subfloat{\includegraphics[width=.2\linewidth]{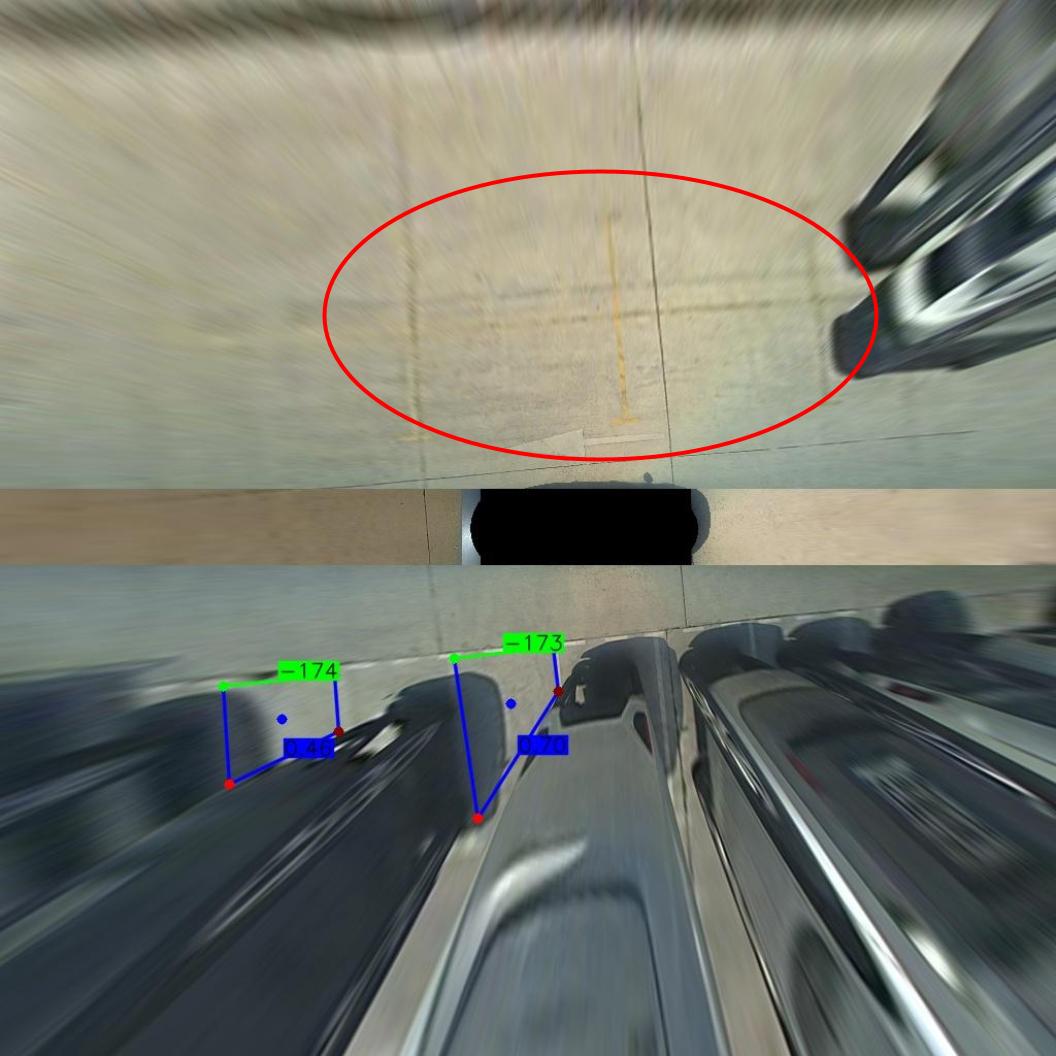}\label{fig:FN2} }
    \qquad
    \subfloat{\includegraphics[width=.2\linewidth]{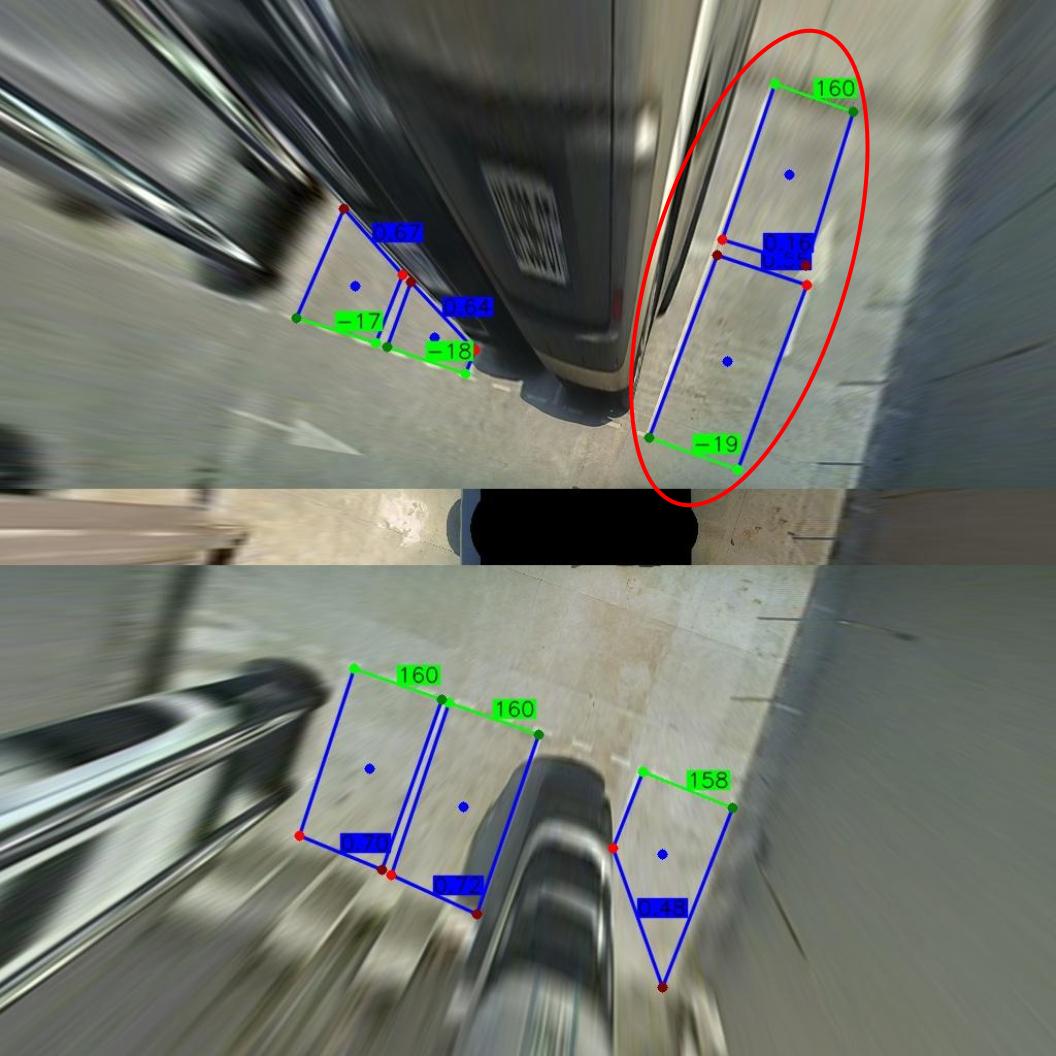}\label{fig:FP1} }
    \qquad  
    \subfloat{\includegraphics[width=.2\linewidth]{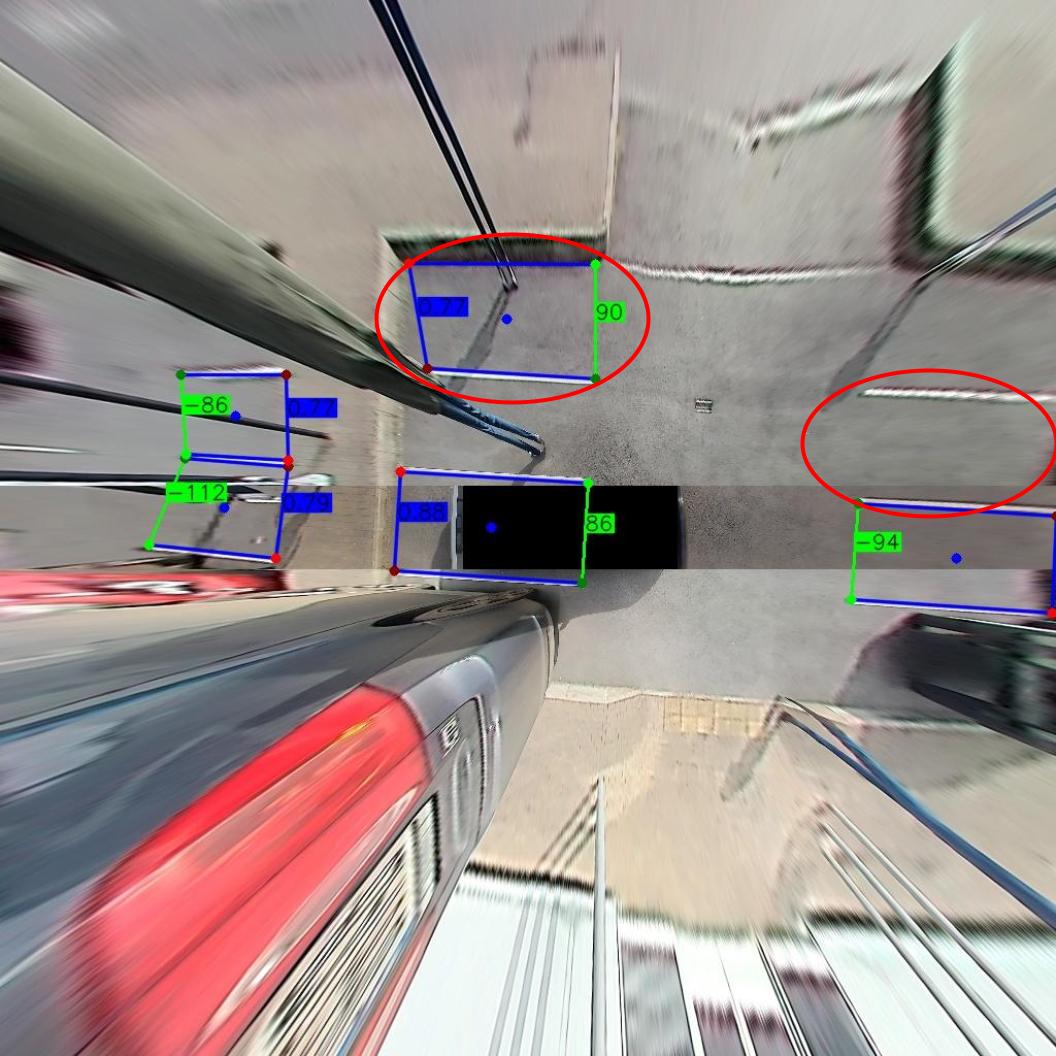}\label{fig:FPFN} }
    \qquad
    \caption{ Some failure cases of detection. False positives and false negatives are marked by red circles.}
    \label{fig:failure_cases}
\end{figure*}

Fig. ~\ref{fig:parking_types} shows some qualitative results of our HPS-Net in diverse scenarios. It is apparent that the algorithm can efficiently detect vacant parking slots around the ego-vehicle, even if they are partially occluded. We also showcase some typical failure cases in ~\ref{fig:failure_cases}. Most failure cases (i.e. false positive or false negative) are caused by image border truncation, blurry markings, or small obstacles present within the parking slot.%Some illustrations from specific use cases are presented as well.

Table~\ref{table:ResultsOnPS20} shows the metrics of our algorithm on the publicly available PS2.0 dataset, which demonstrates the generalization capability of our approach.

Finally, HPS-Net is tested on two different Nvidia embedded platforms: Nvidia Jetson AGX Xavier~\cite{jetson-agx} for general inference purpose and Nvidia Drive AGX Xavier ~\cite{drive-agx} for vehicle integration as mentioned in Section~\ref{section:Methodology:EmbeddedPlatform}. The algorithm with \textit{mish} activation performs at 11 FPS on Nvidia Jetson AGX Xavier (with mish-cuda implementation) while the algorithm with \textit{ReLU} activation performs only at 9 FPS on the same hardware. However, once accelerated with TensorRT, the one with \textit{ReLU} activation achieves 17 FPS on Nvidia Drive AGX Xavier, meeting the real-time requirement in parking scenarios.

\begin{table}
    \caption{\gls{hps-net} results on \gls{vpsd} specific use cases}
    % on a test set of 953 slots from 333 indoor images and 3909 parking slots on 632 paving images
    \begin{center}
    \begin{tabular}{cccccccc}
        \hline
        \textbf{\textit{Use case}}  & \textbf{\textit{Prec.}} & \textbf{\textit{Recall}} & \textbf{\textit{F-1 Sc.}}  & \textbf{\textit{mAP@.5}} & \textbf{\textit{mAP@.5:.95}} % & \textbf{\textit{M. Px. Err.} }
        \\ \hline
        % Indoor & 0.756 & 0.833 & 0.793 & 0.769 & 0.503 \\ \hline
        Indoor & 0.820 & 0.903 & 0.860 & 0.865 & 0.635 \\ \hline %& 2.602 % Mish 1c v5 Base without fine tune.
        % Paving & 0.905 & 0.893 & 0.899 & 0.876 & 0.666 \\ \hline
        Paving & 0.923 & 0.911 & 0.917 & 0.896 & 0.761 \\ \hline % & 2.621 %  Mish 1c v5 Base without fine tune.   
    \end{tabular}
    \end{center}
    \label{table:ResultsOtherTypes}
\end{table}

\begin{table}
    % mish-1c-v5_base fine-tuned on centered PS20
    \caption{Comparison against state-of-the-art methods on public PS2.0 dataset}
    \begin{center}
    \begin{tabular}{ccccccc}
        \hline
        \textbf{\textit{Method}} & \textbf{\textit{Precision}} & \textbf{\textit{Recall}} & \textbf{\textit{F-1 Score}}  & \textbf{\textit{mAP@.5}} \\ \hline
        DeepPS~\cite{Zhang2018} & 0.995 & 0.989 & - & - \\ \hline
        DMPR-PS\cite{Huang2019} & 0.994 & 0.994 & - & - \\ \hline
        AGNN-PD\cite{Min2021} & 0.996 & 0.994 & - & - \\ \hline
        \gls{hps-net} (ours) & 0.998 & 0.999 & 0.998 & 0.999  \\ \hline
    \end{tabular}
    \end{center}
    \label{table:ResultsOnPS20}
\end{table}

\section{CONCLUSIONS}\label{section:Conclusions}

This paper presented \gls{hps-net}, a reliable camera-only holistic parking slot detection approach based on deep learning using surround view fisheye cameras. To address real-time processing and extend the range of detection, a pre-processed \gls{ipm}-based topview image is used, ensuring better orientation, 360° visibility and safe maneuvering. As every centimeter counts when parking, \gls{hps-net} represents each parking slot as a polygon-shaped box instead of a bounding box to accurately predict the position of each visible corner. Entrance line and slot orientation are also detected. \gls{hps-net} demonstrates its superior generalization capacity through open-road testing scenarios with six different demo cars. Furthermore, a combination of the proposed vision solution and ultrasonic mapping could be deployed to manage spaces from simple to complex public installations.

%As every centimeter counts when parking, \gls{hps-net} represents each parking slot as a polygon-shaped box instead of a bounding box to accurately predict the position of each visible corner. Entrance line and slot orientation are also inferred. \gls{hps-net} demonstrates its superior generalization capacity through open-road testing scenarios with different demo cars. Furthermore, a combination of the proposed vision solution and ultrasonic mapping could be deployed to manage spaces from simple to complex public installations (like supermarkets, airports, etc.) for new automated mobility services, such as vehicle sharing, connected and smart city solutions.

Our future work will focus on introducing temporal context with multiple frames for more stable tracked slots, and exploring geometric scene understanding (e.g. visible and invisible extreme points). On the other hand, \gls{ipm} requires precise intrinsic and extrinsic calibration and assumes that the ground is flat. However, in challenging use cases with occlusions or in distant areas, it yields inferior results. Another direction is the implication of recent research works like LSS~\cite{philion2020lift}, CVT~\cite{zhou2022cross}, Simple-BEV~\cite{harley2022simple}, LaRa~\cite{bartoccioni2022lara} that investigate view transformation from image to \gls{bev} with pixel-wise depth estimation, transformers and 3D volume of coordinates over bilinear sampling respectively. Finally, the \gls{vpsd} database will be extended to achieve a more balanced distribution of parking types and to cover more specific use cases, including occupied slots. Last but not least, further work should include a real 3D annotation method, instead of creating ground truths from image or topview domains (i.e. from Lidar or from the prior knowledge of the parking area in form of HD maps combined with precise 3D location of the vehicle in that area).

%\addtolength{\textheight}{-12cm}   % This command serves to balance the column lengths
                                  % on the last page of the document manually. It shortens
                                  % the textheight of the last page by a suitable amount.
                                  % This command does not take effect until the next page
                                  % so it should come on the page before the last. Make
                                  % sure that you do not shorten the textheight too much.

%%%%%%%%%%%%%%%%%%%%%%%%%%%%%%%%%%%%%%%%%%%%%%%%%%%%%%%%%%%%%%%%%%%%%%%%%%%%%%%%

%%%%%%%%%%%%%%%%%%%%%%%%%%%%%%%%%%%%%%%%%%%%%%%%%%%%%%%%%%%%%%%%%%%%%%%%%%%%%%%%

%%%%%%%%%%%%%%%%%%%%%%%%%%%%%%%%%%%%%%%%%%%%%%%%%%%%%%%%%%%%%%%%%%%%%%%%%%%%%%%%
% \section*{APPENDIX}

% Appendixes should appear before the acknowledgment.

\section*{ACKNOWLEDGMENT}

We would like to express our gratitude to Dr. Patrick Pérez, Valeo VP of AI, for his valuable and insightful review of this paper. We also would like to thank our colleagues Thibault Buhet, Sonia Khatchadourian, and Jean-François Duguey for their kind support on data collection and annotation.

%%%%%%%%%%%%%%%%%%%%%%%%%%%%%%%%%%%%%%%%%%%%%%%%%%%%%%%%%%%%%%%%%%%%%%%%%%%%%%%%

\end{document}